\documentclass{article}

\usepackage{microtype}
\usepackage{graphicx}
\usepackage{subcaption}
\usepackage{booktabs} 

\usepackage{hyperref}

\usepackage{graphicx} 

\usepackage[preprint]{icml2026}

\usepackage{amsmath}
\usepackage{amssymb}
\usepackage{mathtools}
\usepackage{amsthm}

\usepackage{multirow}

\usepackage[capitalize,noabbrev]{cleveref}

\theoremstyle{plain}

\theoremstyle{definition}

\theoremstyle{remark}

\usepackage[textsize=tiny]{todonotes}

\icmltitlerunning{Unsupervised Layer-Wise Dynamic Test Time Adaptation for LLMs}

\begin{document}

\twocolumn[
  \icmltitle{Unsupervised Layer-Wise Dynamic\\
  Test Time Adaptation for LLMs}



  \icmlsetsymbol{equal}{*}

  \begin{icmlauthorlist}
    \icmlauthor{Longhuan Xu}{seu-monash}
    \icmlauthor{Cunjian Chen}{monash}
    \icmlauthor{Feng Yin}{seu}
  
  \end{icmlauthorlist}

  \icmlaffiliation{seu-monash}{Southeast University-Monash University Joint Graduate School}
  \icmlaffiliation{seu}{Southeast University}
  \icmlaffiliation{monash}{Monash University}
  
  \icmlcorrespondingauthor{Feng Yin}{yinfeng@seu.edu.cn}
  
  \icmlkeywords{Machine Learning, ICML}

  \vskip 0.3in
]



\printAffiliationsAndNotice{}  

\begin{abstract}
Test-time adaptation (TTA) for large language models (LLMs) updates model parameters at inference time using signals available at deployment. This paper focuses on a common yet under-explored regime: \textbf{unsupervised, sample-specific TTA}, where the model adapts independently for each prompt using only the prompt itself, without gold answers or external supervision. Although appealing, na\"ive unsupervised TTA with a fixed, handcrafted learning rate can be unstable: updates may overfit to prompt-specific statistics, drift from the desired answer distribution, and ultimately degrade generation quality. This failure mode is not surprising, as in this case TTA must adapt to a single prompt within only a few gradient steps, unlike standard training that averages updates over large datasets and long optimization horizons. Therefore, we propose \textbf{layer-wise dynamic test-time adaptation}, a framework which explicitly modulates TTA strength as a function of prompt representation, LLM structure and adaptation step. In our setting, TTA updates only LoRA parameters, and a lightweight hypernetwork predicts \textbf{per-layer, per-step learning-rate multipliers}, enabling fine-grained control. Experiments across various datasets and LLMs consistently show that our method substantially strengthens TTA by learning effective scaling patterns over adaptation steps and transformer layer projections, improving stability while delivering better performance.
\end{abstract}

\begin{figure*}
  \vskip 0.2in
  \begin{center}
    \centerline{\includegraphics[width=0.95\textwidth]{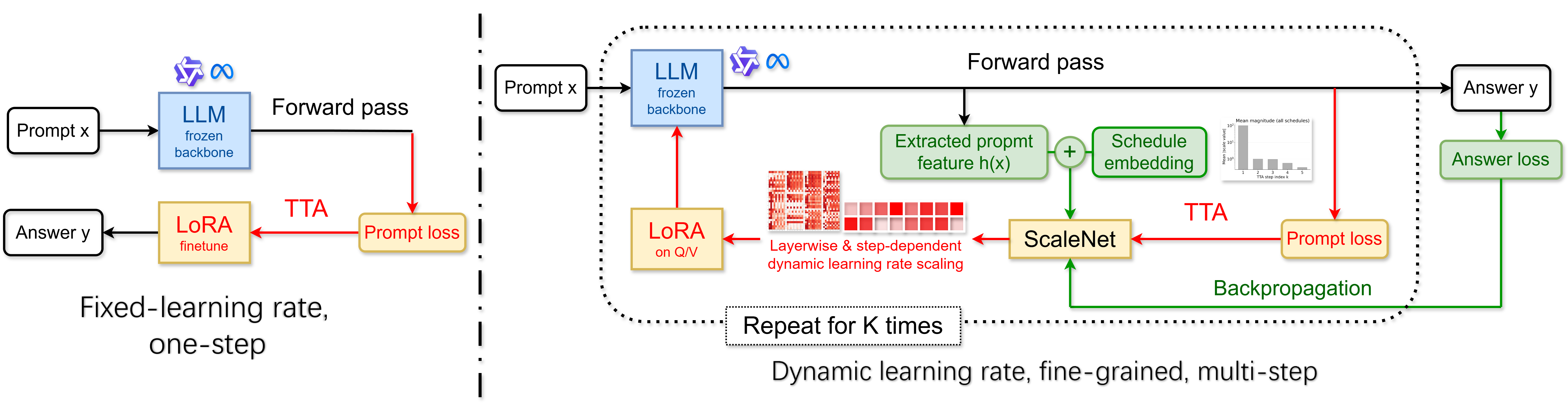}}
    \caption{
      Unsupervised layer-wise dynamic TTA training pipeline (right side).}
    \label{fig:pipeline}
    \vspace{-2mm}
  \end{center}
  \vspace{-2mm}
\end{figure*}

\section{Introduction}
Large language models (LLMs) have become a general-purpose backbone for natural language understanding and generation, powering applications ranging from open-ended assistants to domain-specific writing and reasoning tools \cite{chowdhery2022palmscalinglanguagemodeling, singh2025openaigpt5card}. Despite strong overall performance, its real-world usage often differs from the training regime: prompts vary wildly in style, length and content, domain-specific terminology appears frequently, and user instructions can deviate from patterns seen during pretraining and alignment. This motivates test-time optimization \cite{sun2020testtimetrainingselfsupervisiongeneralization} methods that adjust model behavior to bridge the gap between deployment context and the training corpus.

A second, orthogonal challenge for LLMs is instance-level objective mismatch \cite{krause2019dynamicevaluationtransformerlanguage, rannentriki2024revisitingdynamicevaluationonline}. Standard training minimizes the expected (or average) loss over the entire training set, producing a single parameter setting that is necessarily a global compromise across heterogeneous samples. At test time, however, performance is determined by the loss on the current prompt instance. Even when the test distribution matches training in aggregate, the globally optimal parameters need not be optimal for a particular prompt, leaving room for per-instance specialization.

During inference, LLM behavior can be adjusted in two different ways.
First, \emph{prompting-based} methods—such as in-context learning \cite{dong2024surveyincontextlearning} and retrieval-augmented generation (RAG) \cite{lewis2021retrievalaugmentedgenerationknowledgeintensivenlp}—steer LLM by modifying the input context while keeping model parameters frozen. Second, \emph{test-time adaptation} (TTA) \cite{hu2025testtimelearninglargelanguage} modifies the model itself by performing parameter updates during inference. A practically dominant deployment regime is \emph{unsupervised, sample-specific} TTA. ``\textbf{Unsupervised}" means that we typically observe only the prompt $x$ and rarely have access to the gold continuation/answer $y$ (or any other supervision). ``\textbf{Sample-specific}" means per-instance updates: for each prompt $x$, the model takes a small number of gradient steps on $x$ itself (e.g., minimizing prompt negative log-likelihood), then generates $y$ and resets before the next query. This adapt-and-reset protocol matches real-world usage, where prompt streams are heterogeneous and often non-stationary; otherwise, offline (or continual) fine-tuning would be more appropriate. Unsupervised, sample-specific adaptation is also attractive operationally because it requires no labels or external data, can be applied on-the-fly, and can leverage parameter updates to encode information beyond the finite context window length of LLMs \cite{dai2019transformerxlattentivelanguagemodels}.

However, na\"ive unsupervised TTA which runs a few gradient steps on the prompt \cite{hu2025slotsamplespecificlanguagemodel} with a fixed learning rate is often brittle for LLMs. Because the optimization signal comes solely from the observed prompt, updates can over-amplify prompt-specific statistics and deviate from the desired continuation/answer distribution, degrading generation quality. This procedure is also highly sensitive to adaptation strength: small learning rates produce negligible improvements within a handful of steps, whereas large learning rates can cause destructive parameter changes. More fundamentally, sample-specific TTA is driven by a single instance, so its gradients have high-variance and are easily dominated by idiosyncratic prompt features. In our experiments, effective adaptation calls for an aggressive first update (often unevenly distributed across layers) followed by rapid damping, and the appropriate scale can even depend on the LoRA initialization. In summary, a single fixed global learning rate cannot accommodate these step- and layer-dependent dynamics.

In this paper we propose the \textbf{unsupervised layer-wise dynamic TTA} framework aimed at controlling the most direct and influential hyperparameters in TTA: the learning rate. The central idea is that the appropriate adaptation strength should depend on both the input prompt and the adaptation state, and that a single global TTA learning rate is too coarse for deep transformers. Gradient magnitudes and update sensitivity can vary substantially across layers and across TTA steps; as a result, uniform updates may be either too small to matter or large enough to be harmful. We therefore introduce a learnable, fine-grained lightweight scaling hypernetwork, called \textsc{ScaleNet}, which predicts per-layer, per-step learning rate multipliers for test-time updates. Intuitively, it serves as a control mechanism: it suppresses potentially harmful updates in sensitive layers or for delicate stages, while allowing stronger adaptation where it is beneficial.

In our implementation, test-time adaptation updates only LoRA \cite{hu2021loralowrankadaptationlarge} parameters in the attention query/value projections while keeping the pretrained backbone frozen; before TTA backpropagation, \textsc{ScaleNet} predicts multipliers that rescale the base learning rate for LoRA.

Our method can also be viewed through the lens of test-time scaling \cite{muennighoff2025s1simpletesttimescaling, zeng2025revisitingtesttimescalingo1like}. Rather than improving inference by increasing sampling, reranking candidates, or performing multi-pass generation, our framework scales adaptation compute: it uses a small budget of test-time gradient steps and learns how to allocate update magnitudes across layers and steps on a per-sample basis.

Our contributions can be summarized as:
\begin{itemize}
  \item \textbf{Layer-wise dynamic learning rate for unsupervised, sample-specific test time adaptation (TTA).}
  We propose a dynamic TTA scheme that replaces a single global learning rate with learned, step- and layer-dependent learning rates, improving stability and effectiveness under a small adaptation budget.

  \item \textbf{Lightweight per-layer (Q/V) scaling framework.}
  A hypernetwork predicts non-negative multipliers for each transformer layer (separately for query/value LoRA matrices) and each TTA step, enabling fine-grained control of update magnitudes during test-time learning.

  \item \textbf{Efficient training via first-order approximation.}
  We train the framework by unrolling the same TTA procedure used at inference and optimizing it with a first-order approximation that avoids expensive second-order derivatives.

\end{itemize}

\section{Related Works}
\subsection{Large Language Models}
Large language models (LLMs) have become the dominant paradigm in modern NLP. They are typically transformer-based \cite{vaswani2023attentionneed} auto-regressive decoder models trained with next-token prediction, and scaling model size, data, and compute tends to improve generalization and unlock emergent capabilities, including zero-/few-shot in-context learning and multi-step chain-of-thought (COT) \cite{wei2023chainofthoughtpromptingelicitsreasoning} reasoning. In practice, LLMs also undergo post-training to better match human intent. Instruction tuning improves instruction following, while alignment methods such as reinforcement learning from human feedback (RLHF) \cite{ouyang2022traininglanguagemodelsfollow} shape outputs toward helpfulness and safety.

Alongside these training recipes, architectural and system advances further improve performance. Mixture-of-experts (MoE) \cite{fedus2022switchtransformersscalingtrillion} increases model capacity with sparse activation and keeps per-token compute relatively low. Retrieval-augmented generation (RAG) grounds generation in external knowledge, improving factuality and enabling maintenance through an editable retrieval index rather than weight updates. More recently, LLM agents couple an LLM with planning, memory, and tool/API execution loops, allowing it to decompose goals into actionable steps and carry out workflows beyond passive prompting.

\subsection{Test-Time Adaptation}
Test-time adaptation (TTA) updates a deployed model at inference time using unlabeled test-time signals. A canonical early approach is \textsc{TENT}, which performs unsupervised adaptation by minimizing prediction entropy on target batches, updating only lightweight affine parameters (and normalization statistics) \citep{wang2021tentfullytesttimeadaptation}. Since na\"ive entropy minimization can cause overconfident drift or even collapse, conservative variants such as \textsc{COME} stabilize adaptation by modeling uncertainty and optimizing a guarded surrogate \citep{zhang2024cometesttimeadaptionconservatively}. 

For autoregressive LLMs, recent work argues that entropy is often misaligned with generation, and that perplexity (next-token negative log-likelihood) is a more suitable self-supervised objective for test-time updates \citep{hu2025testtimelearninglargelanguage}. Moreover, sample-specific (per-prompt) adaptation has been shown feasible, e.g., \textsc{SLOT} performs a few optimization steps for each prompt under an adapt-and-reset protocol \citep{hu2025slotsamplespecificlanguagemodel}.

\subsection{Low-Rank Finetuning}
Parameter-efficient fine-tuning (PEFT) adapts large pretrained models by updating only a small set of additional parameters while keeping the backbone frozen. A widely used PEFT method is Low-Rank Adaptation (LoRA) \cite{hu2021loralowrankadaptationlarge}, which represents weight updates with low-rank factors, greatly reducing trainable parameters and optimization cost. LoRA is commonly applied to transformer attention projections (e.g., query/value), and its updates can be merged into the base weights for deployment with negligible inference overhead.

\section{Proposed Method}

\subsection{Unsupervised Sample-Specific TTA for LLMs}
Suppose an LLM parameterized by $\Theta$ is trained on a data distribution $p(x)$:
\begin{equation}
LLM_\Theta \;=\; \arg\min_{\Theta}\ \mathbb{E}_{x\sim p(x)}\big[-\log P(x;\Theta)\big].
\end{equation}
At test time, LLM $P_\Theta$ will be evaluated on a downstream distribution $q(x,y)$, from which a test-time prompt $x$ and  the desired continuation/answer $y$ are sampled. By Bayes' rule, the conditional distribution $q(y\mid x)$ is determined by both the joint $q(x,y)$ and the marginal $q(x)$:
\begin{equation}
q(y\mid x) \;=\; \frac{q(x,y)}{q(x)} \;\propto\; q(x)\ \text{under }\mathcal{C}(x).
\label{eq:bayes}
\end{equation}
In the unsupervised setting where $y$ is unavailable, we seek a control method such that, under appropriate constraints $\mathcal{C}(x)$,  adapting the model using only the observed prompt $x$ can still improve answer prediction. Intuitively, prompts and answers are often semantically aligned, so optimizing on $x$ can provide a useful learning signal for getting LLM prediction $P_{\Theta}(y\mid x)$ closer to ground truth $q(y\mid x)$.

Following \cite{hu2025testtimelearninglargelanguage}, we consider unsupervised test-time adaptation using the standard negative log-likelihood objective to update LLM parameterized by ${\Theta}$. Backpropagation can be written as:
\begin{equation}
\Theta' \;=\; \Theta - \eta \nabla_{\Theta}\!\Big(-\log P(x;\Theta)\Big).
\label{eq:backprop}
\end{equation}
Here $\Theta'$ denotes the new parameters after one prompt-only update on $x$, and $\eta$ is learning rate. A first-order Taylor expansion of new conditional log probability $\log P_{\Theta'}(y\mid x)$ using Eq.~\eqref{eq:backprop} yields:
\begin{equation}
\log P_{\Theta'}(y\mid x)
=
\log P_{\Theta}(y\mid x)
+
\eta\langle g_x,\ g_y\rangle
+
\mathcal{O}(\eta^2).
\label{eq:taylor}
\end{equation}
Here the cross-gradient term $\langle g_x,\ g_y\rangle$ is defined as the inner product between prompt gradient $g_x=\nabla_{\Theta}\log P(x;\Theta)$ and answer gradient
$g_y=\nabla_{\Theta}\log P_{\Theta}(y\mid x)$ conditioned on the same prompt:
\begin{equation}
\langle g_x,\ g_y\rangle 
\;:=\;
\Big\langle \nabla_{\Theta}\log P(x;\Theta),\ \nabla_{\Theta}\log P_{\Theta}(y\mid x)\Big\rangle .
\end{equation}
Our goal is to design a control mechanism that tries to guarantee $\langle g_x, g_y\rangle \ge 0$, so that unsupervised TTA tends to increase the conditional log-likelihood, i.e., $\log P_{\Theta'}(y\mid x) \ge \log P_{\Theta}(y\mid x)$.

In general, TTA can be viewed as a special case of fine-tuning performed at inference time. Updating all LLM parameters on a single prompt is not only computationally prohibitive, but also conceptually inappropriate, as it can easily cause severe overfitting. We therefore adopt the widely used LoRA \cite{hu2021loralowrankadaptationlarge} parameterization:
\begin{equation}
W' \;=\; W + \Delta W,\quad 
\Delta W \;=\; B A.
\label{eq:lora}
\end{equation}
where $W\in\mathbb{R}^{d_{\text{out}}\times d_{\text{in}}}$ is a frozen weight matrix, and $A\in\mathbb{R}^{r\times d_{\text{in}}}$, $B\in\mathbb{R}^{d_{\text{out}}\times r}$ are trainable low-rank factors with rank $r\ll \min(d_{\text{out}},d_{\text{in}})$. Given all trainable LoRA parameters $\Phi$, Eq.~\eqref{eq:backprop} becomes:
\begin{equation}
\Phi' \;=\; \Phi - \eta \nabla_{\Phi}\!\Big(-\log P(x;\Phi)\Big).
\end{equation}

Because prompts can vary drastically across instances, we study sample-specific test-time adaptation, treating each prompt as an independent adaptation episode. Concretely, for each test prompt $x$, we start with a fresh set of LoRA factors by re-initializing the low-rank matrix $\Delta W = BA$ to zero (e.g. $B=0$ and $A$ random), run a small number of gradient steps, generate the response $y$ with the adapted parameters, and then discard the adaptation state. This \emph{adapt-and-reset} protocol separates unrelated queries and matches real-world interactive usage.

\begin{figure}
  \vskip 0.2in
  \begin{center}
    \centerline{\includegraphics[width=1.0\columnwidth]{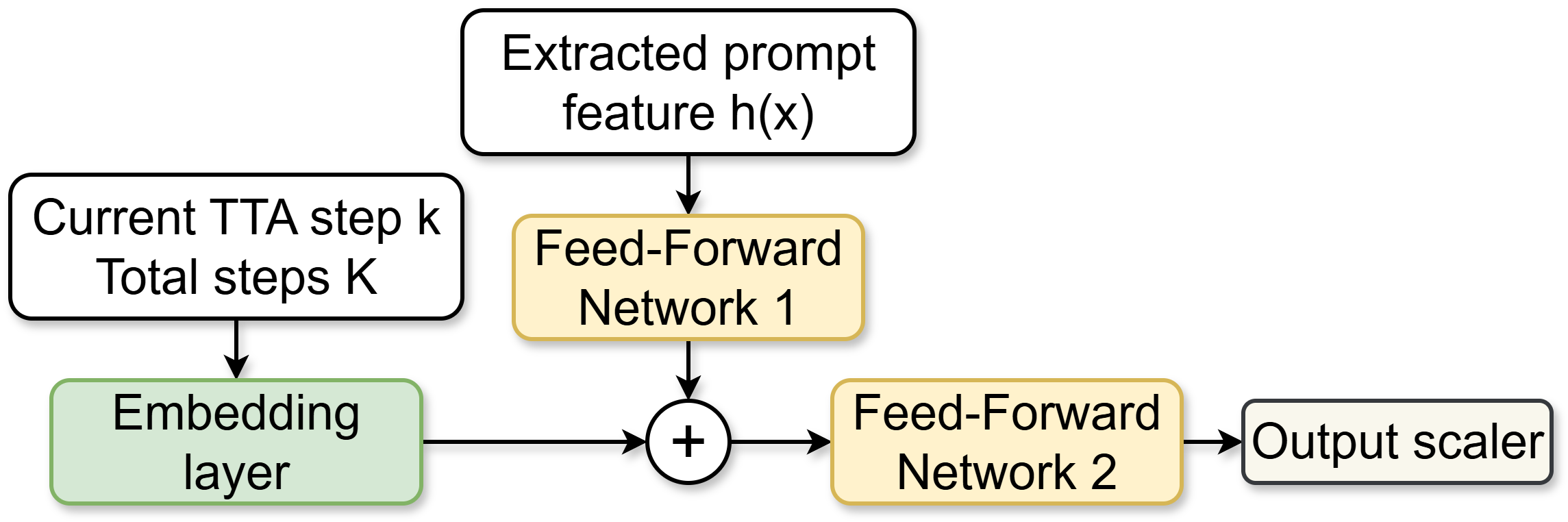}}
    \caption{
      Simple control hypernetwork: \textsc{ScaleNet} architecture.}
    \label{fig:ScaleNet}
    \vspace{-2mm}
  \end{center}
  \vspace{-2mm}
\end{figure}

\subsection{Layer-wise Hypernetwork Control}
How to find the correct constraint $\mathcal{C}(x)$? Eq.~\eqref{eq:bayes} exposes a fundamental challenge of unsupervised TTA: adapting on $x$ inevitably increases the model’s probability mass on the observed prompt, effectively boosting the marginal term $q(x)$. Such an increase improves the conditional probability of $y$ given $x$ only if the joint probability $q(x,y)$ increases \emph{even more}—that is, the update must amplify $q(x,y)$ beyond the marginal gain on $q(x)$:
\begin{equation}
\frac{q'(x,y)}{q(x,y)} \;>\; \frac{q'(x)}{q(x)}.
\label{eq:joint_vs_marginal_gain}
\end{equation}
where $q'(x,y)$ and $q'(x)$ are the joint and marginal distributions after TTA. In the absence of gold answers, it is difficult to handcraft constraints that reliably enforce this condition.

Nevertheless, prompts and answers are not independent: paired $(x,y)$ are inherently coupled through semantics and task structure. It is thus natural to infer useful properties of $(x,y)$ from $x$ alone, and neural networks are well suited to capture these hard-to-specify relationships. This idea remains consistent with our unsupervised TTA setting: the gold answer is used only during training to teach a neural network how to infer useful properties of $(x,y)$ from $x$, and it is never accessed during real test-time adaptation.

Moreover, revisiting the next-token prediction objective shows that, although the unsupervised loss is written as $\log P(x;\Theta)$, it is actually computed token-wise, i.e., a collection of conditional prediction tasks rather than a ``true'' probability of the prompt as a whole \cite{radford2019gpt2}:
\begin{equation}
\sum_t \log P(x_t \mid x_{<t};\Theta)
\end{equation}
Therefore, unsupervised TTA primarily updates the model to better match conditional distributions under the context induced by $x$, and it may be reasonable to expect that these conditional distributions can transfer from observed prompt tokens to unseen answer tokens by increasing:
\begin{equation}
\log P_{\Theta}(y\mid x)
\approx
\sum_{t}\log P_{\Theta}\!\left(y_t \mid x, y_{<t}\right).
\label{eq:pyx_factorization}
\end{equation}

To make it clear, we introduce a hypernetwork $\mathcal{H}_{\psi}(x)$ parameterized by $\psi$ that always takes $x$ as input to produce the constraint $\mathcal{C}$ from the prompt:
\begin{equation}
\mathcal{C}(x) \;=\; \mathcal{H}_{\psi}(x).
\label{eq:coarse}
\end{equation}
The training objective of hypernetwork would be to find an optimum solution $\psi^\star$ minimizing answer loss after TTA:
\begin{equation}
\psi^\star
\;=\;
\arg\min_{\psi}\ 
\mathbb{E}_{(x,y)\sim q(x,y)}
\Big[
f\big(\Theta'(x,\mathcal{C}(x)),\, y\big)
\Big].
\end{equation}
Here $\Theta'(x,\mathcal{C}(x))$ denotes the parameters of LLM after applying the unsupervised TTA (e.g., one or multiple gradient steps on $x$) under constraint $\mathcal{C}(x)$, starting from the base parameters $\Theta$; $f\big(\Theta'(x,\mathcal{C}),\, y\big)$ denotes the LLM loss referenced on gold answer.

Among all the constraints in TTA, \textbf{we choose ``learning rate", one of the most straightforward yet critical control signals, as the to-be-optimized hypernetwork output $\mathcal{C}$}. Considering that TTA typically allows only a handful of gradient steps, it often requires a learning rate far larger than standard fine-tuning; otherwise, with a conventional rate (e.g., $10^{-5}$) and only a few updates (e.g., three), the resulting parameter change is effectively negligible. While the number of TTA steps also affects performance, it is usually a less trainable hyper-parameter: empirically, gains increase with additional steps but saturate quickly.

In this paper, we focus on transformer-based LLMs, which remain the dominant architecture in current practice and are composed of many stacked transformer layers \cite{bommasani2022opportunitiesrisksfoundationmodels}. We propose that a single global TTA learning rate is a coarse (or even bad) control knob because gradient characteristics and update sensitivity can vary substantially across layers and steps, and that a multiplicative layer-wise scaling of the learning rate at each TTA step is a better optimizable constraint for achieving larger TTA gains while maintaining stability. 

Formally speaking, let $L$ denote the number of layers and $k\in\{1,\dots,K\}$ the current TTA step out of a planned total of $K$ steps. Constraint $\mathcal{C}$ is now defined as a layer-wise learning-rate scaler predicted at each TTA step:
\begin{equation}
\mathcal{C}=\;\{s^{(k)}\}_{k=1}^{K}, 
\quad 
s^{(k)}=\mathcal{H}_{\psi}(x,k,K)\in\mathbb{R}_{\ge 0}^{L}.
\label{eq:C_sequence}
\end{equation}

Trainable LoRA parameters $\Phi$ are then decomposed into $\Phi=\{\phi_\ell\}_{\ell=1}^L$ ($\ell$ is layer index) and with the base learning rate $\eta$, our layer-wise dynamic TTA update at layer $\ell$, step $k$ reads:
\begin{equation}
\phi_\ell^{(k+1)}
=
\phi_\ell^{(k)}
-\eta\, s_\ell^{(k)} \,\nabla_{\phi_\ell}\!\Big(-\log P(x;\Phi^{(k)})\Big).
\label{eq:layerwise_tta}
\end{equation}

\begin{figure*}[t]
\vskip 0.0in
\centering
\includegraphics[width=0.225\textwidth,clip,trim=5 0 5 0]{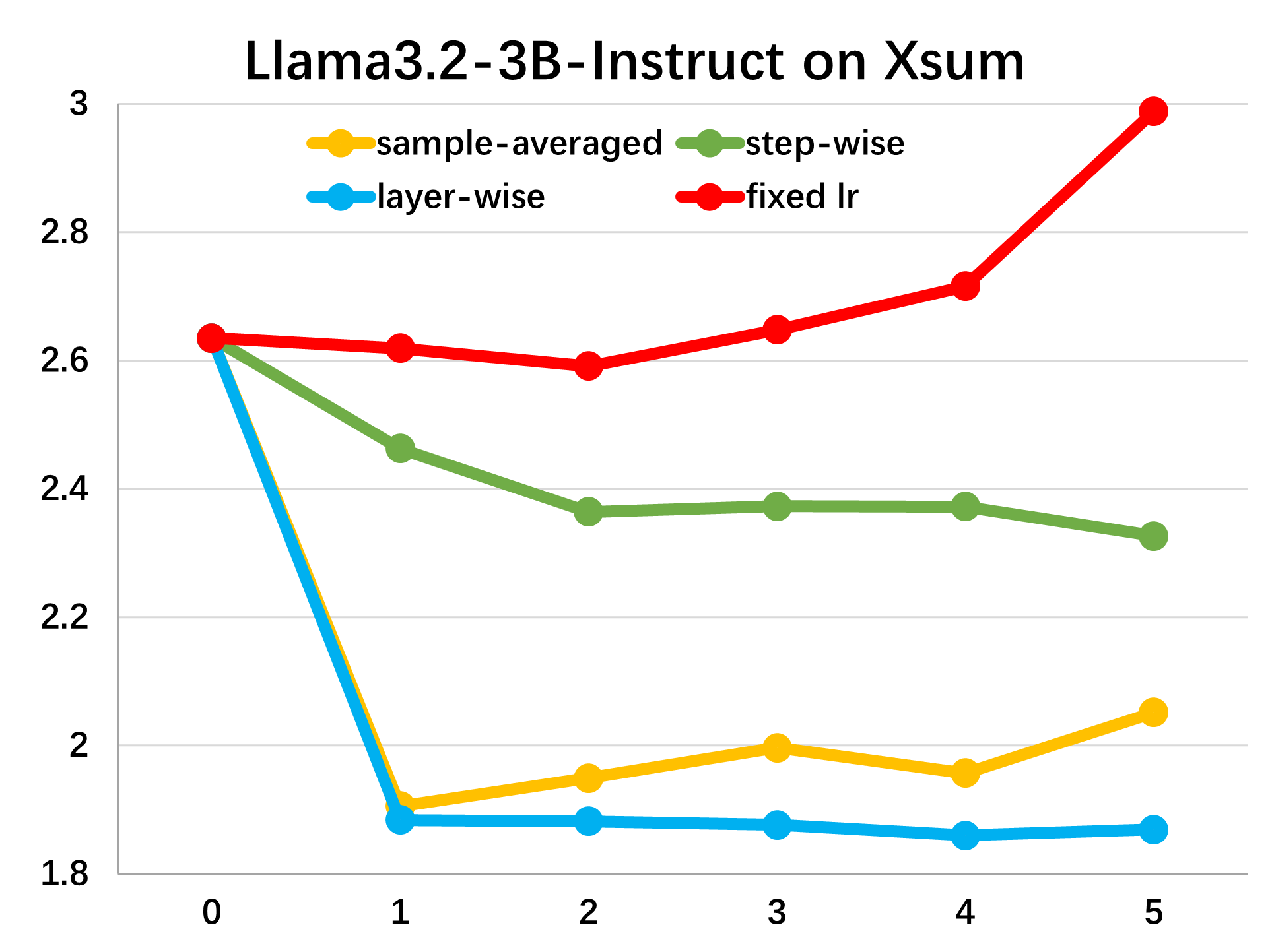}\hfill
\includegraphics[width=0.225\textwidth,clip,trim=5 0 5 0]{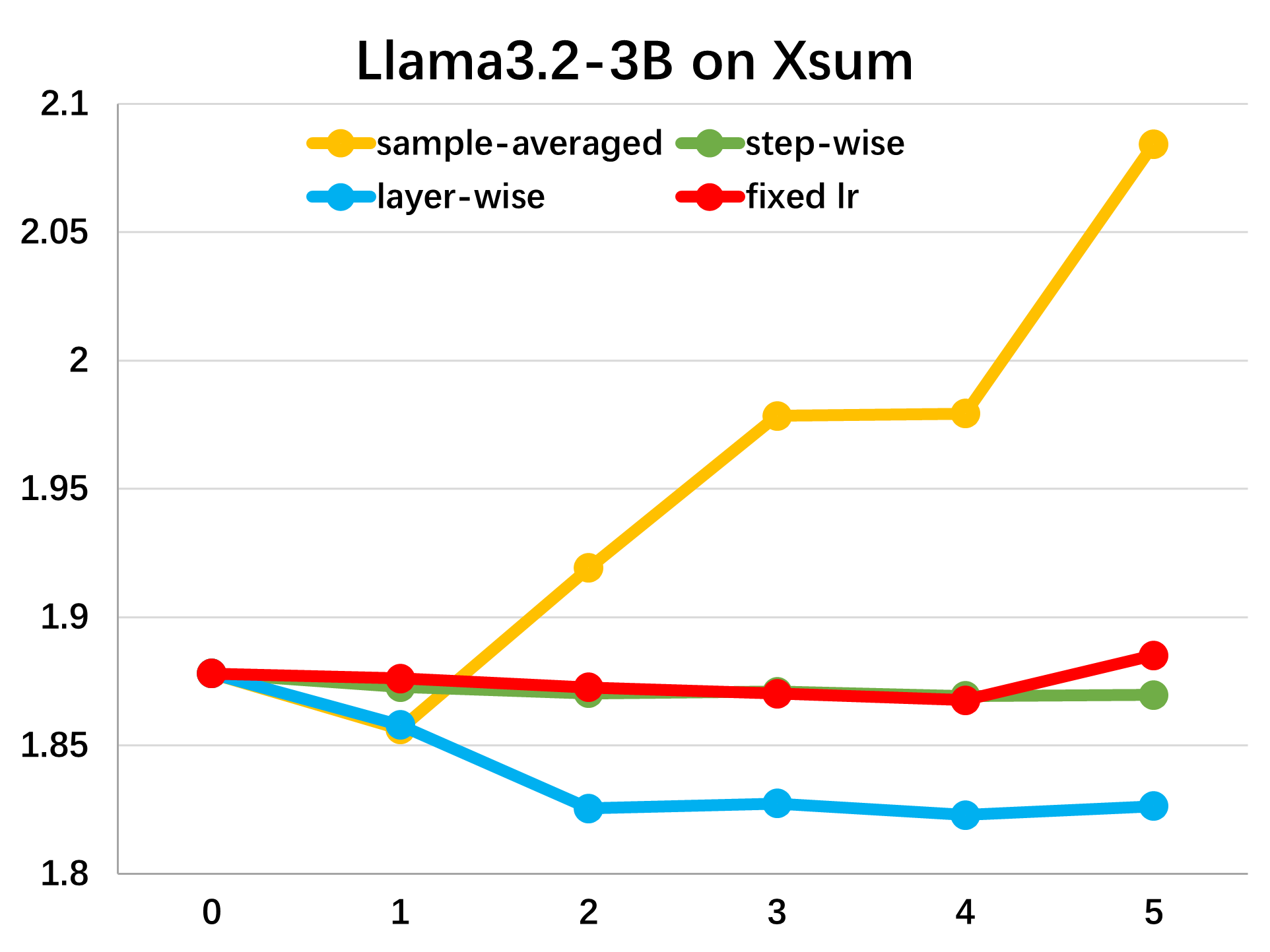}\hfill
\includegraphics[width=0.225\textwidth,clip,trim=5 0 5 0]{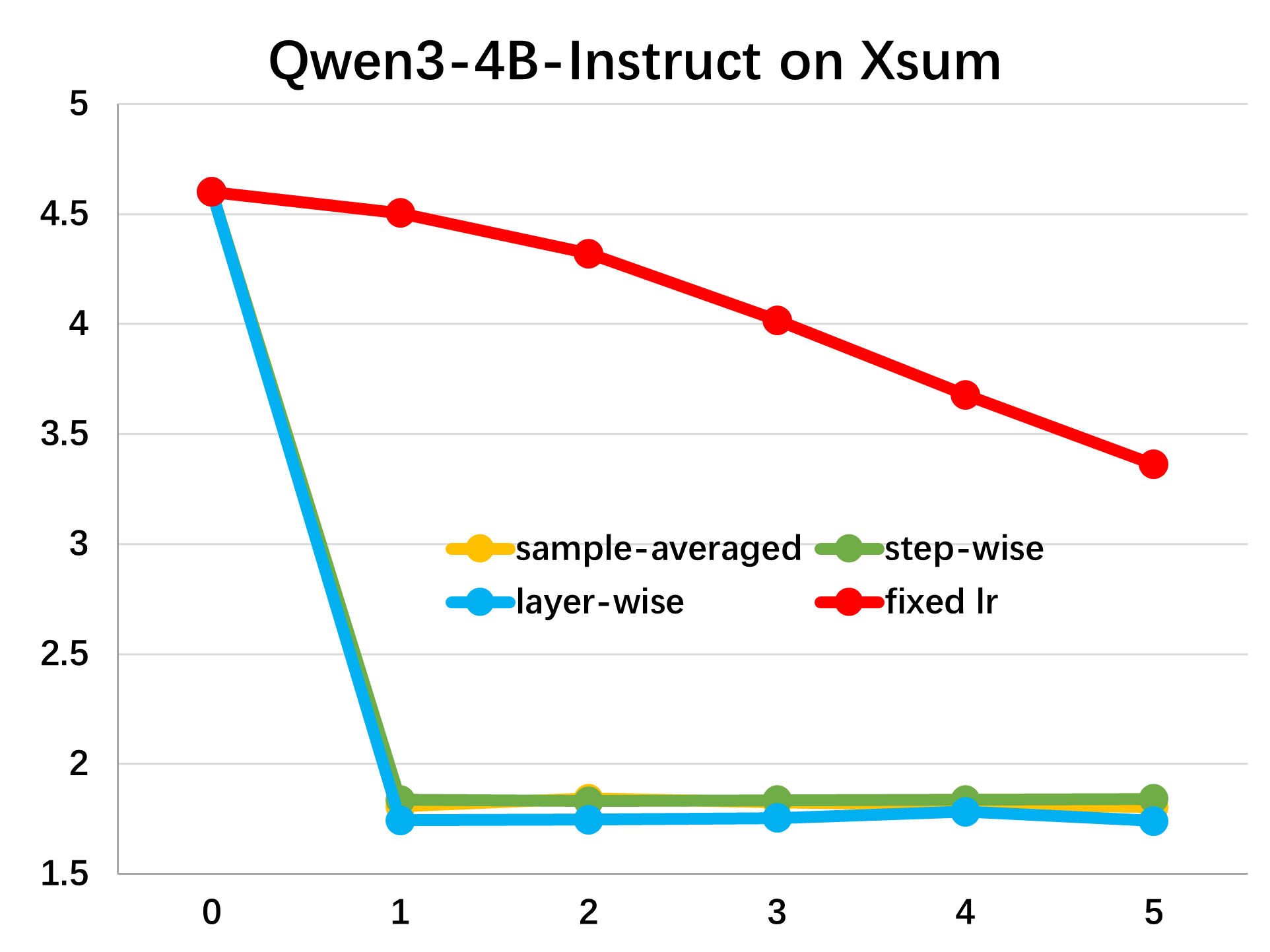}\hfill
\includegraphics[width=0.225\textwidth,clip,trim=5 0 5 0]{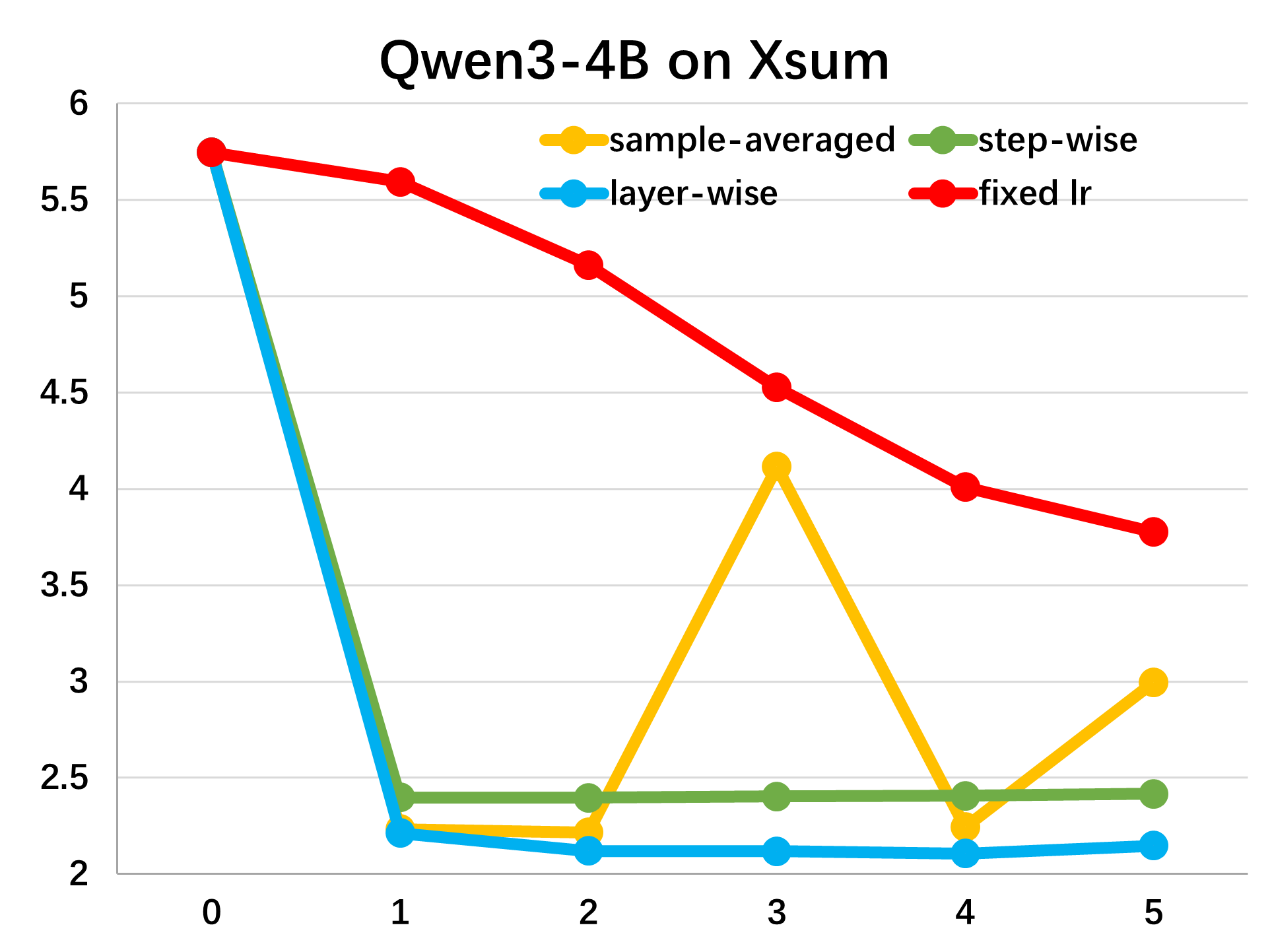}

\includegraphics[width=0.225\textwidth,clip,trim=5 0 5 0]{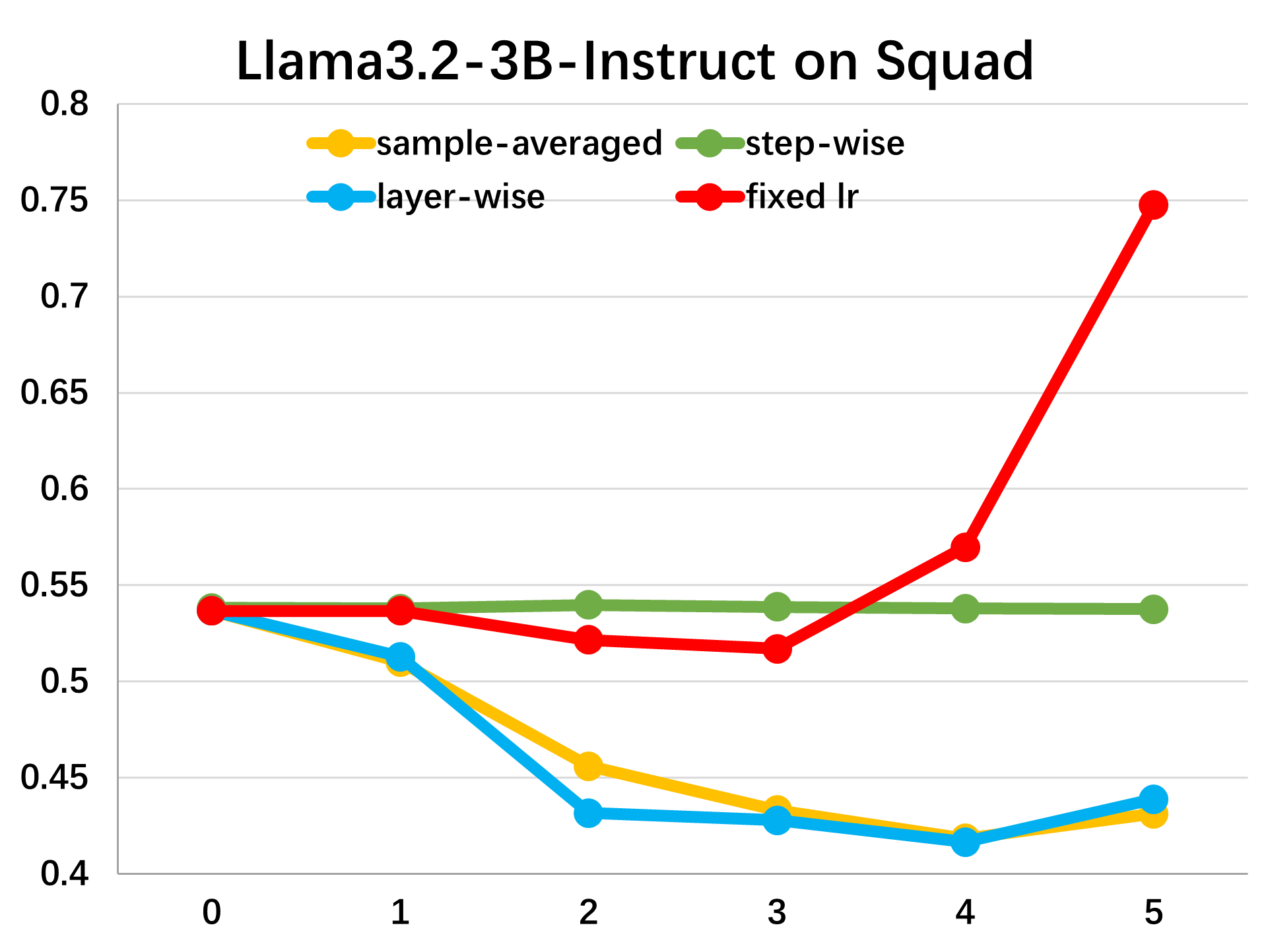}\hfill
\includegraphics[width=0.225\textwidth,clip,trim=5 0 5 0]{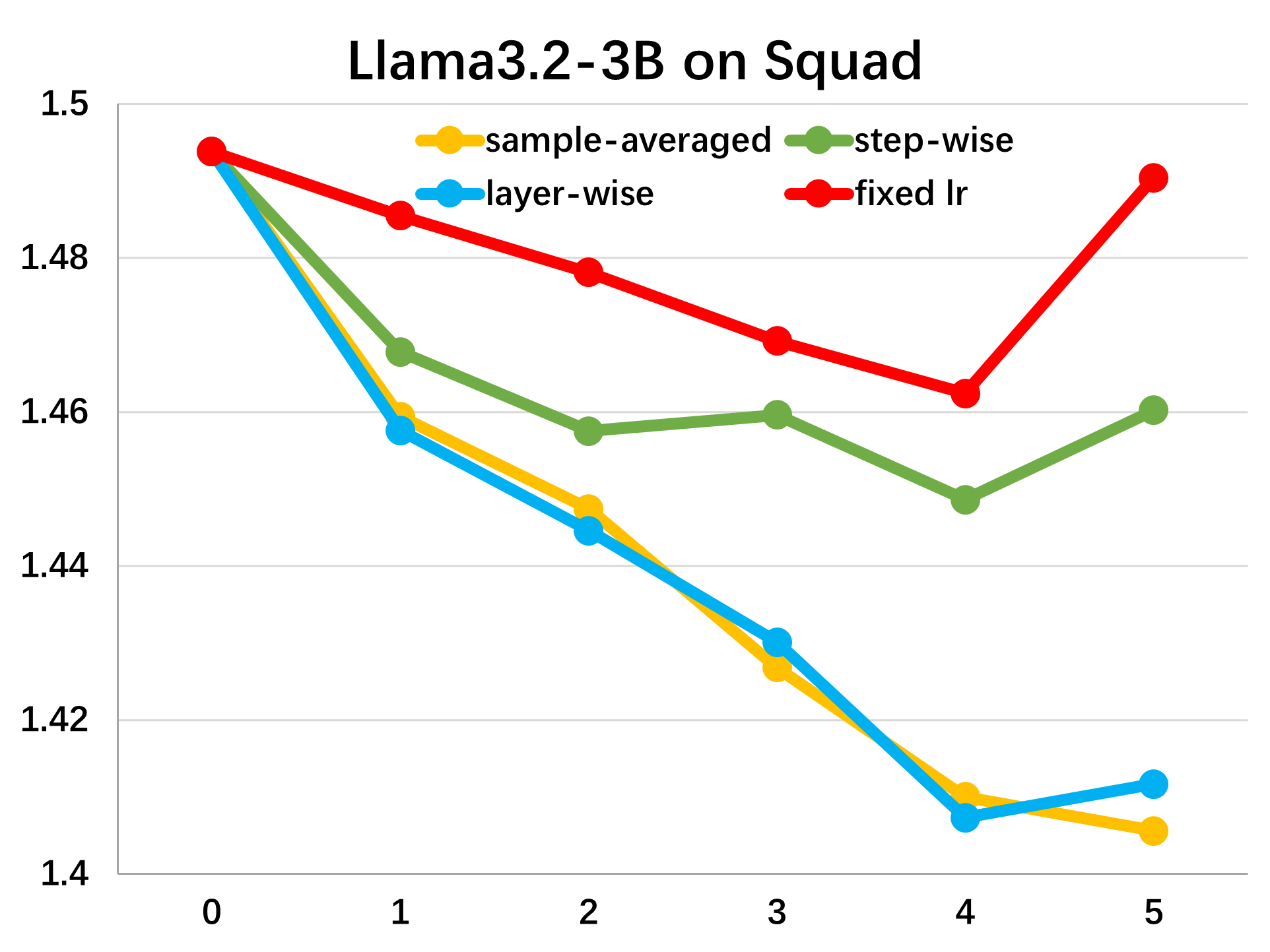}\hfill
\includegraphics[width=0.225\textwidth,clip,trim=5 0 5 0]{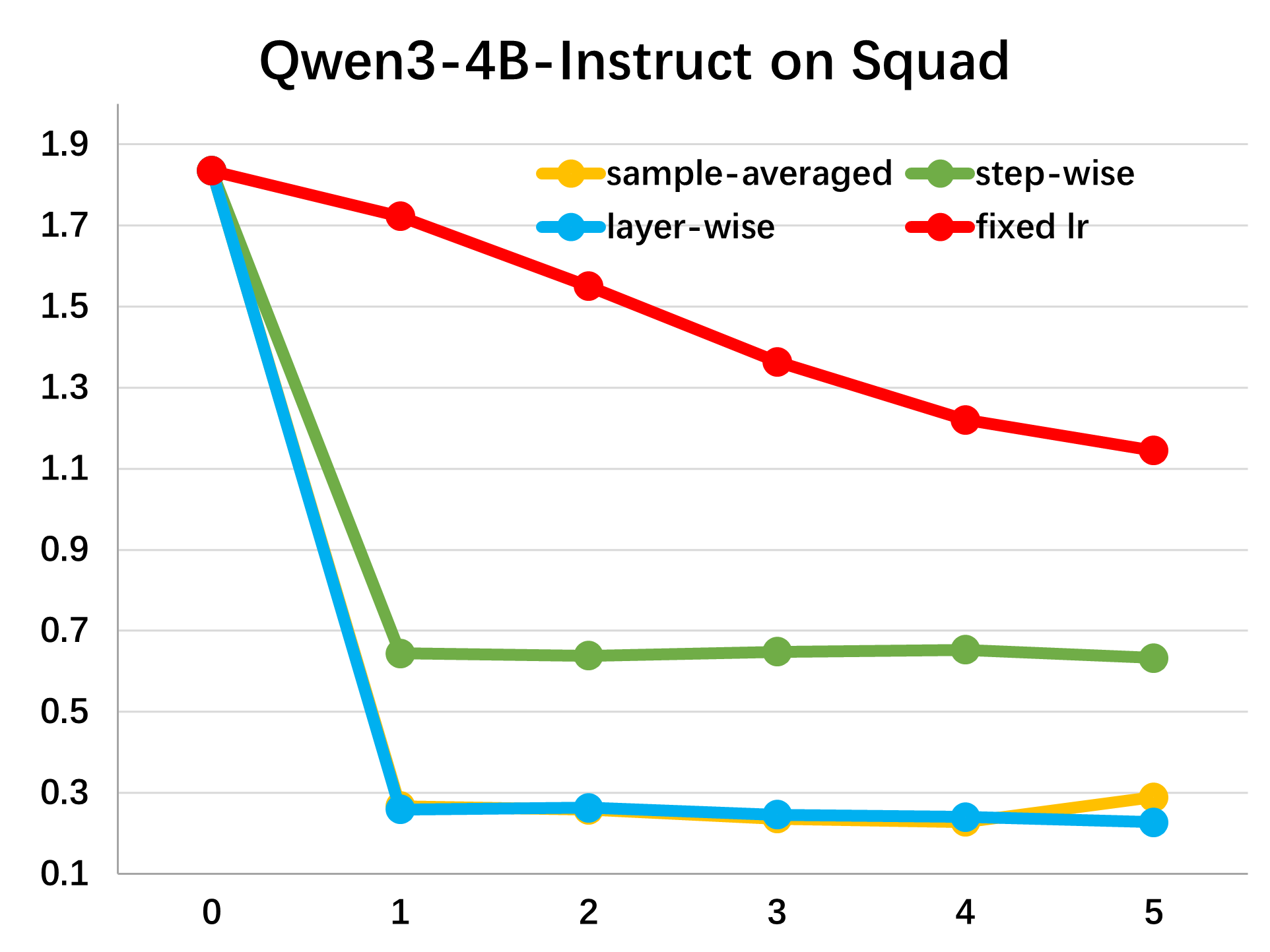}\hfill
\includegraphics[width=0.225\textwidth,clip,trim=5 0 5 0]{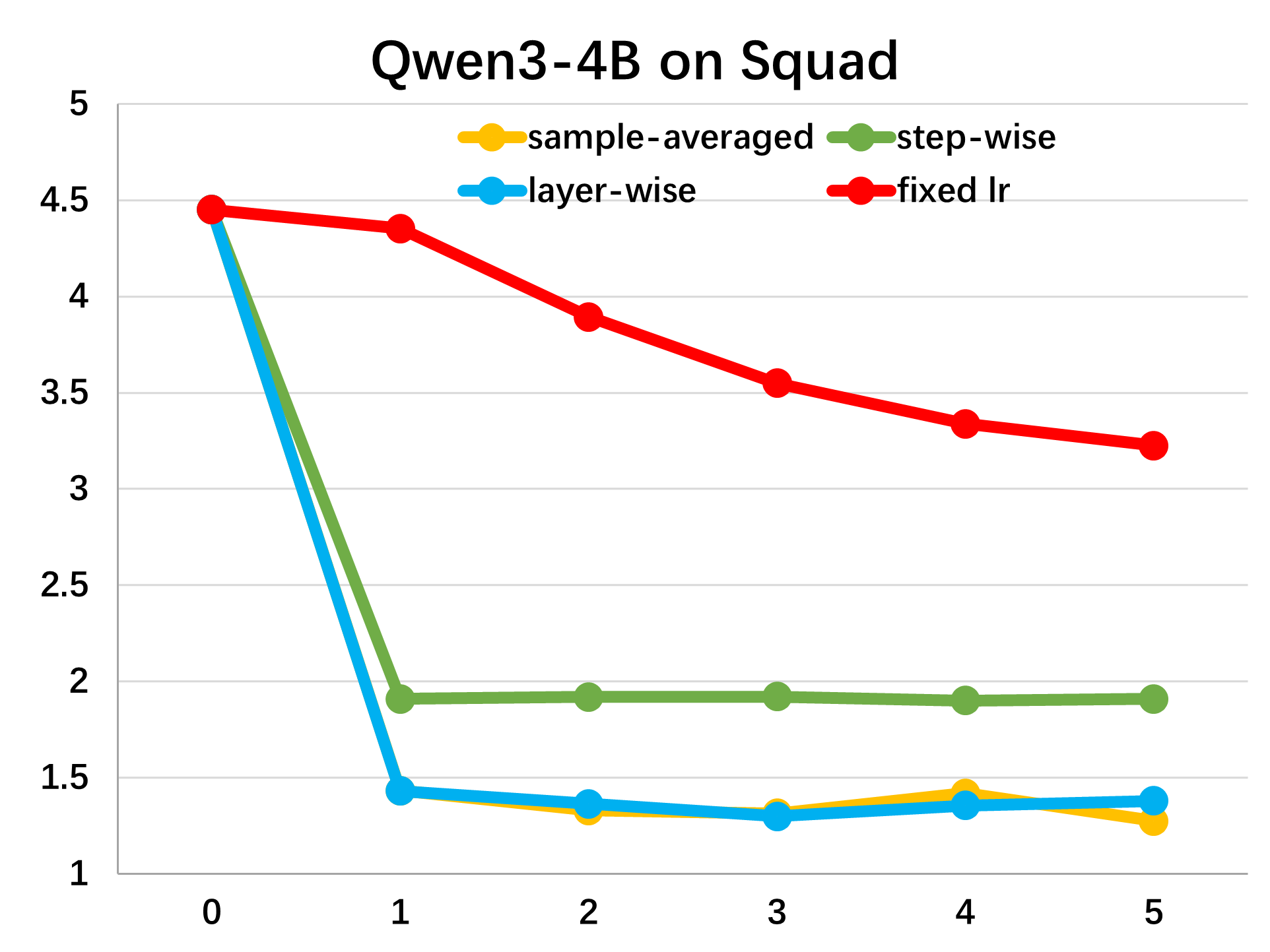}

\includegraphics[width=0.225\textwidth,clip,trim=5 0 5 0]{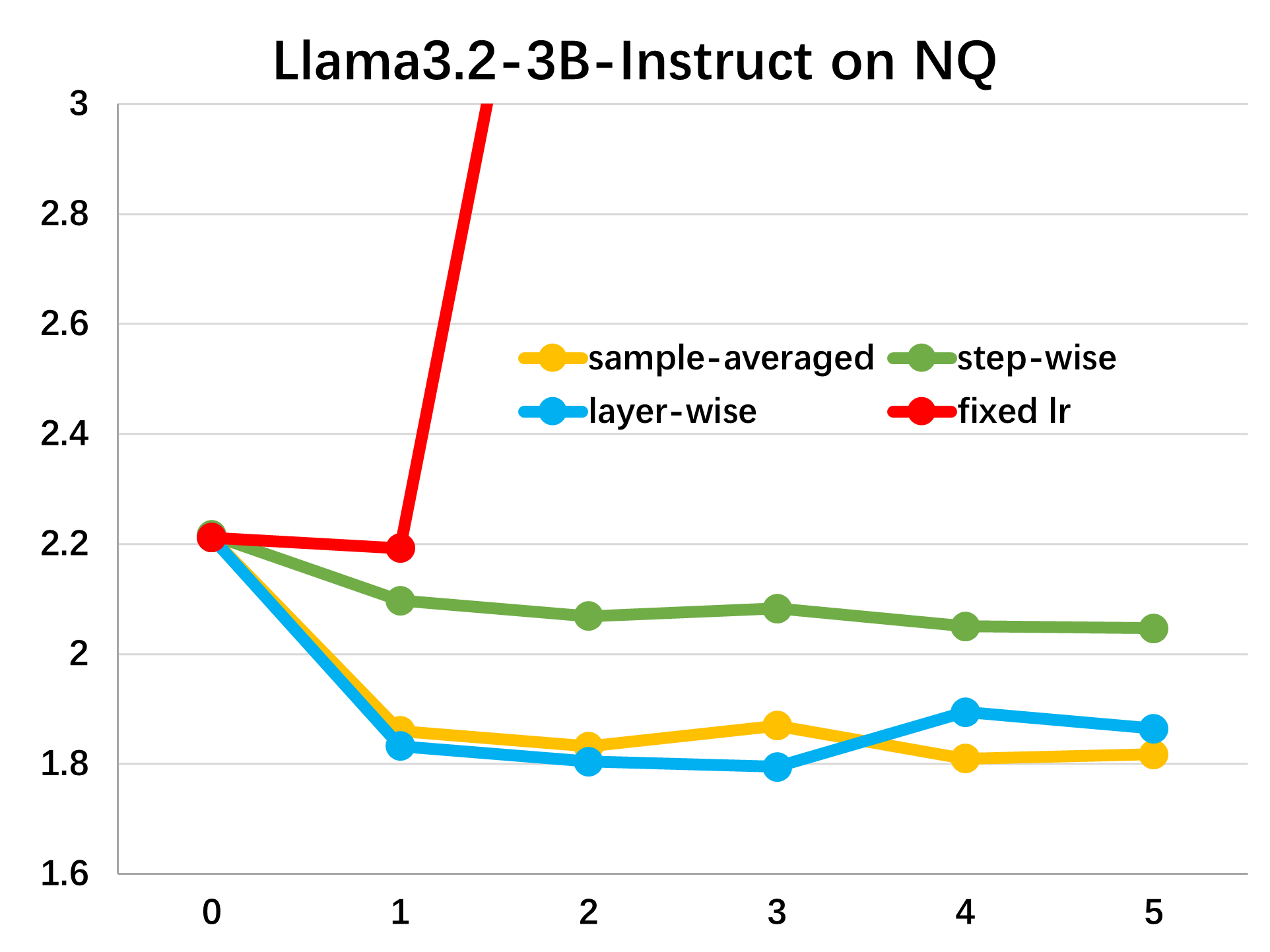}\hfill
\includegraphics[width=0.225\textwidth,clip,trim=5 0 5 0]{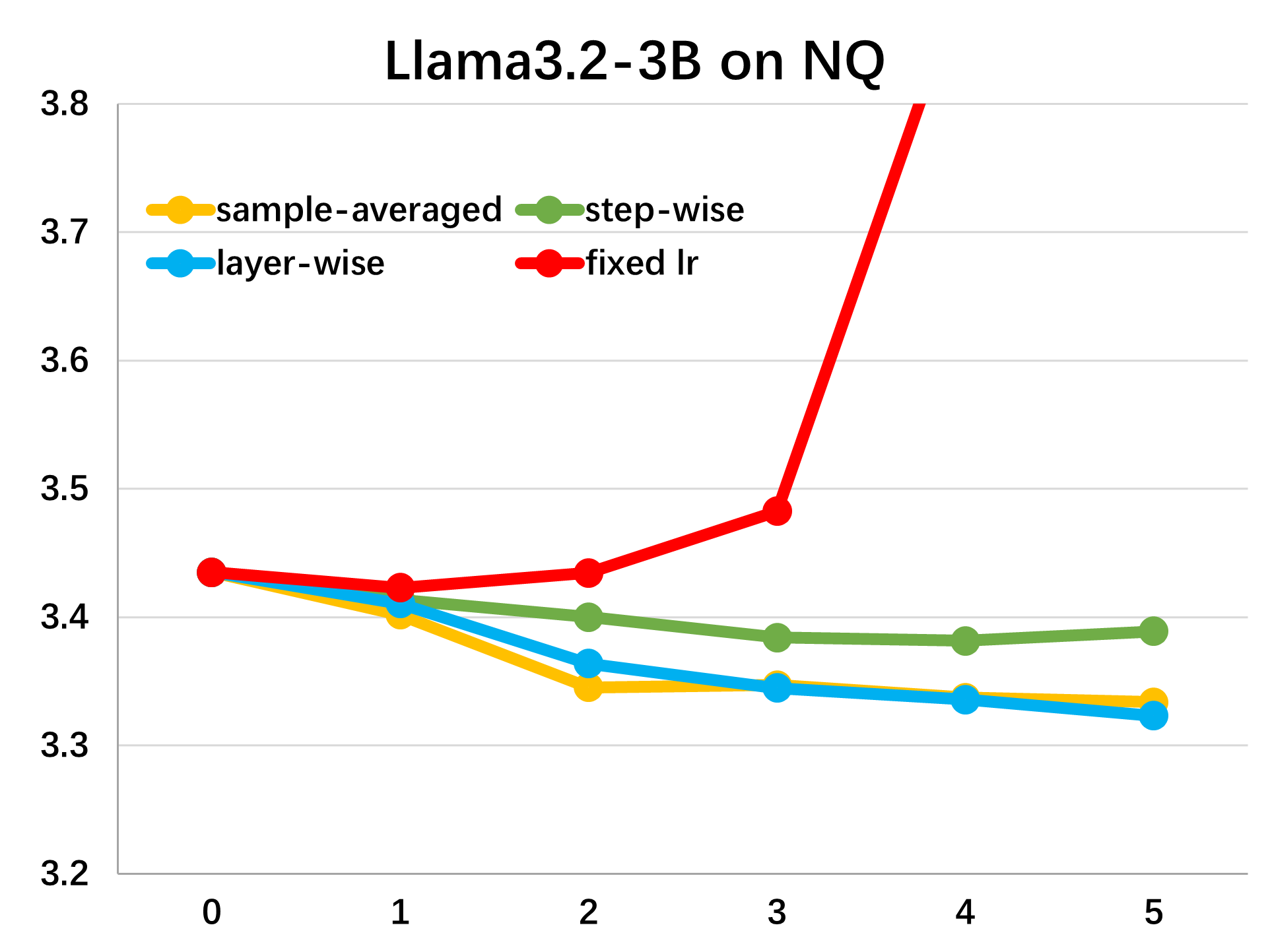}\hfill
\includegraphics[width=0.225\textwidth,clip,trim=5 0 5 0]{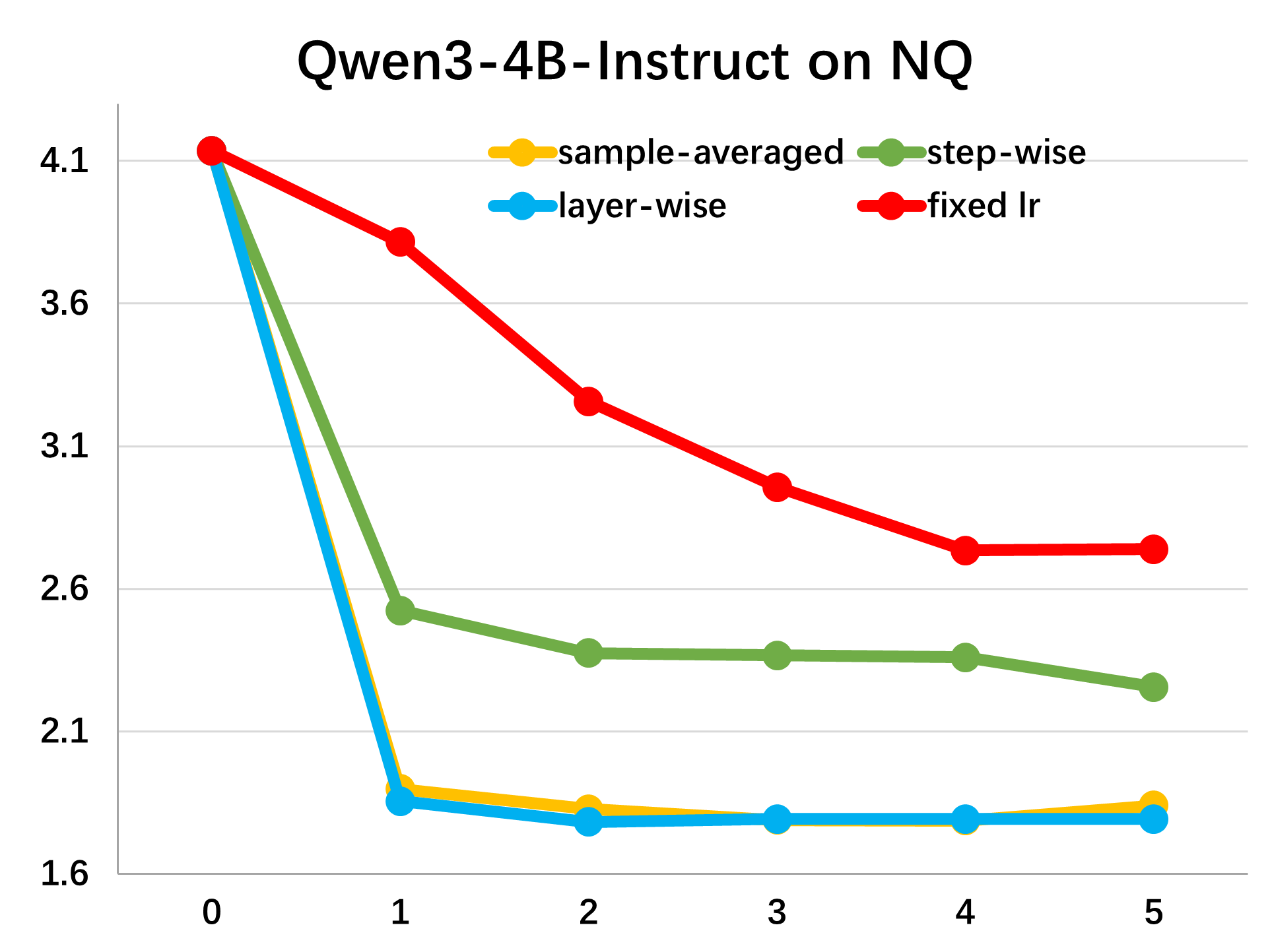}\hfill
\includegraphics[width=0.225\textwidth,clip,trim=5 0 5 0]{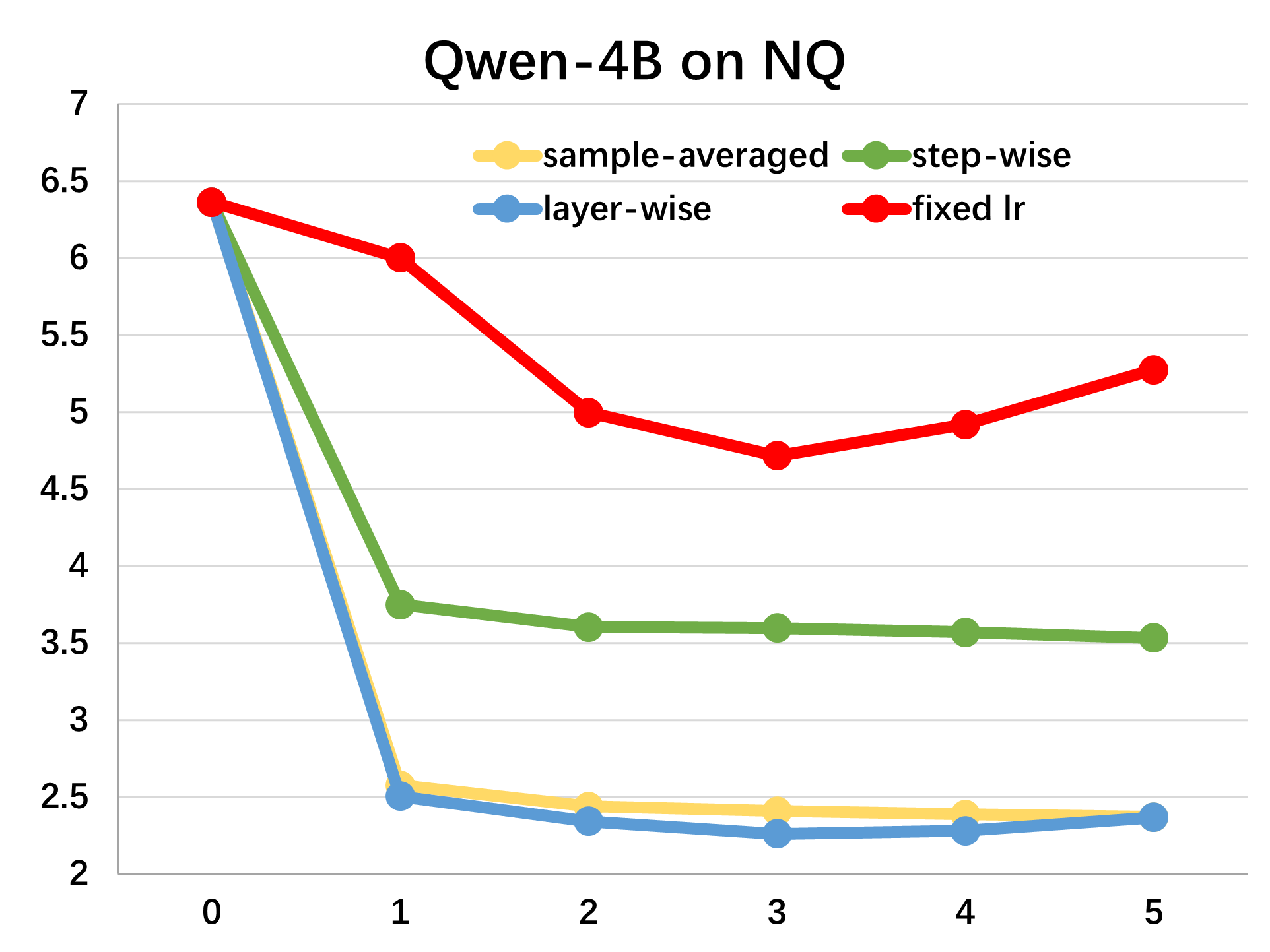}

\includegraphics[width=0.225\textwidth,clip,trim=5 0 5 0]{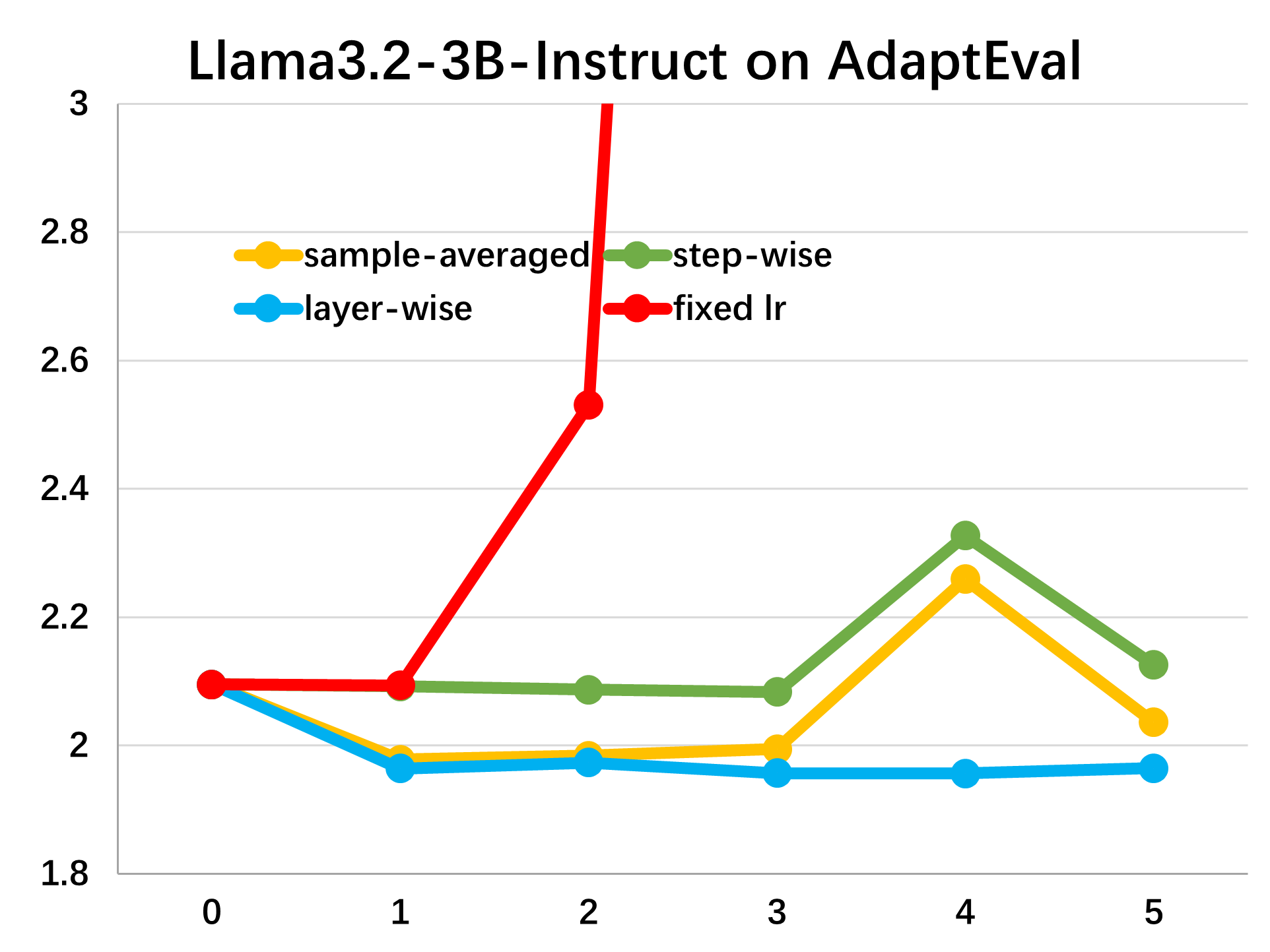}\hfill
\includegraphics[width=0.225\textwidth,clip,trim=5 0 5 0]{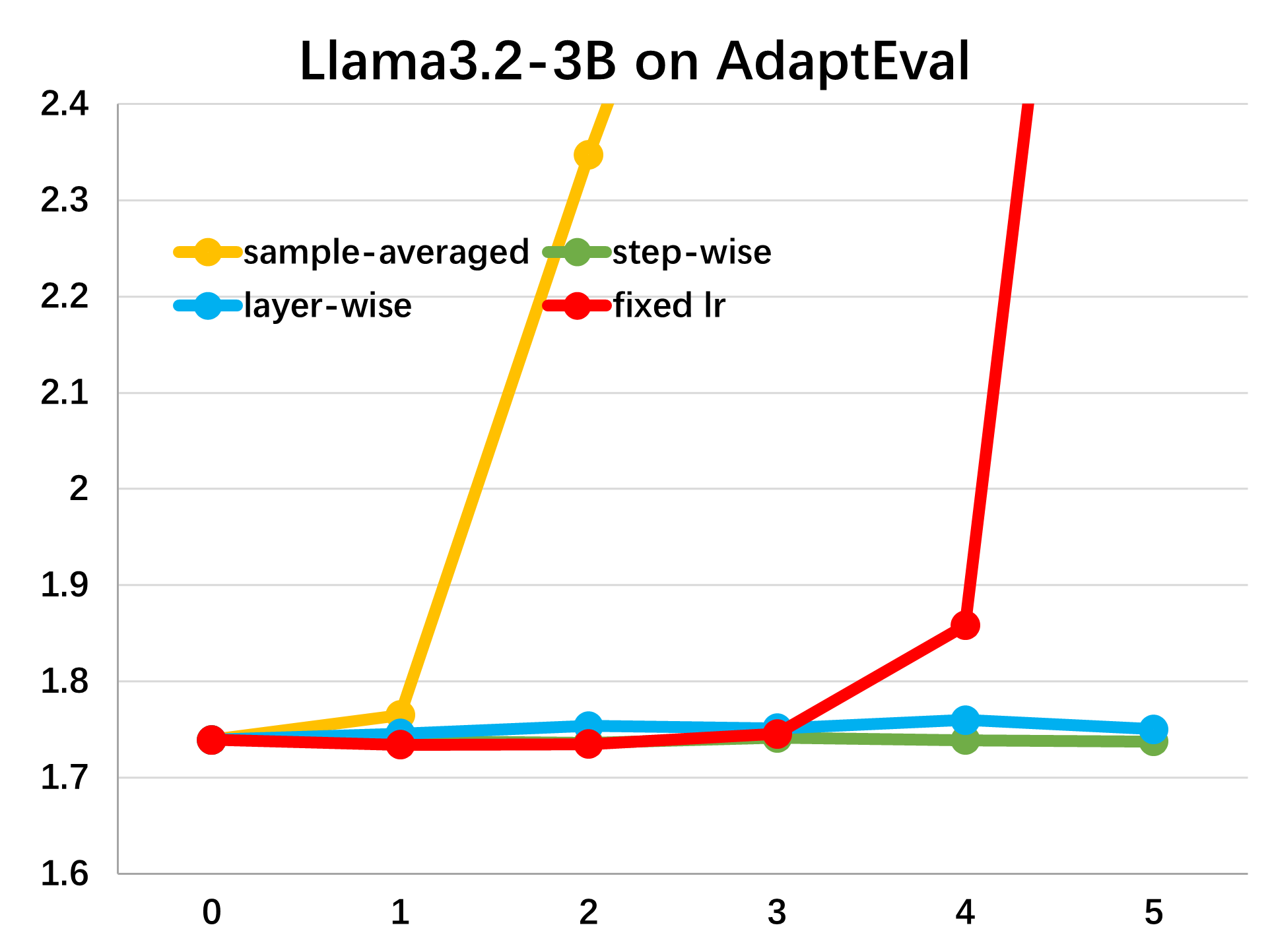}\hfill
\includegraphics[width=0.225\textwidth,clip,trim=5 0 5 0]{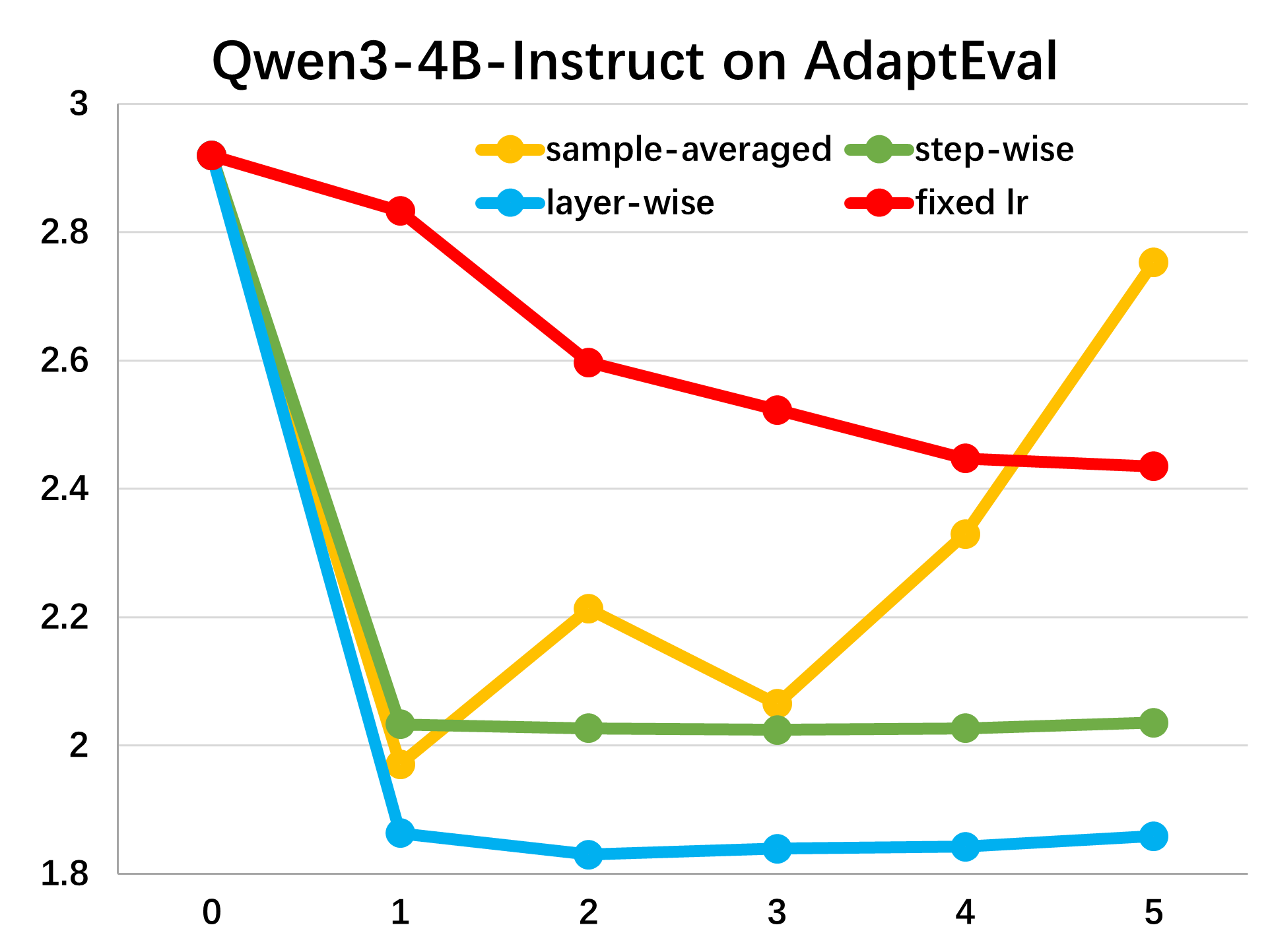}\hfill
\includegraphics[width=0.225\textwidth,clip,trim=5 0 5 0]{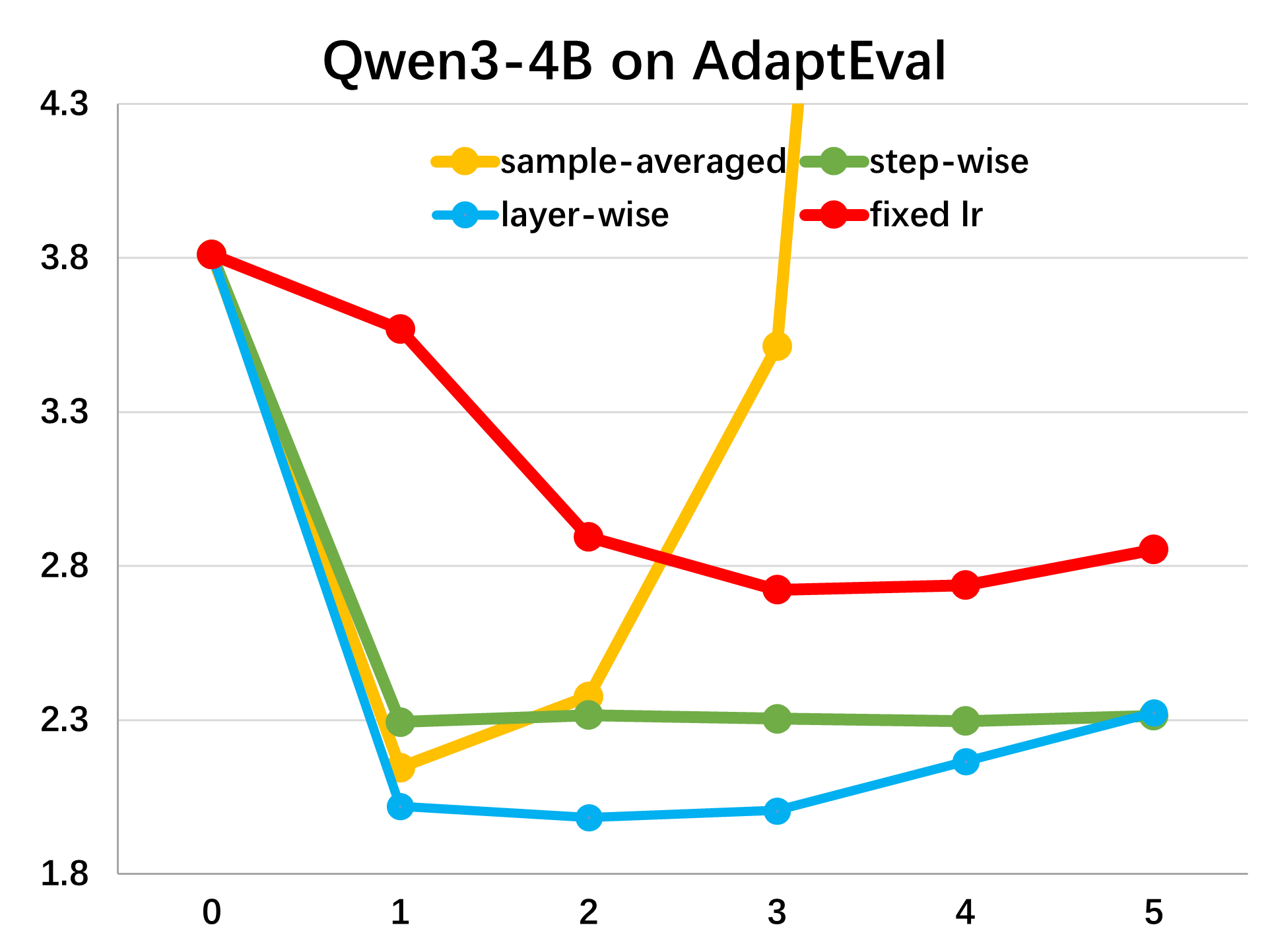}

\vskip 0.0in
\caption{\textbf{NLL results.} The vertical axis shows the average negative log-likelihood (NLL) per answer token, and the horizontal axis shows the number of TTA steps. The red curve is the na\"ive fixed-learning-rate baseline. Green and blue correspond to layer-agnostic/step-wise and layer-wise \textsc{ScaleNet}; yellow corresponds to sample-averaged layer-wise \textsc{ScaleNet}.}
\label{fig:base_LLMs}
\end{figure*}

\subsection{Dynamic Control Framework}

We now describe the hypernetwork $\mathcal{H}_\psi$ named as \textsc{ScaleNet} and its framework in detail. Unsupervised, layer-wise dynamic test time adaptation consists of several stages that alternate in a loop (\cref{fig:pipeline}). First, the LLM performs a forward pass on the prompt $x$. Next, the hypernetwork outputs a dynamic learning-rate scaler. Then, the query and value weight matrices\footnote{Although we refer to the framework as \emph{layer-wise}, we in fact predict two scales per layer---one for the query projection and one for the value projection---to increase control capacity.} in each transformer layer of the LLM undergo LoRA fine-tuning based on the next-token-prediction negative log-likelihood of the prompt, using a learning rate adjusted by the hypernetwork output. This loop is repeated for $K$ times until the scheduled test-time adaptation is finally finished.

As mentioned in Eq.~\eqref{eq:C_sequence}, \textsc{ScaleNet} takes the prompt $x$, the current TTA step $k$, and the total scheduled number of TTA steps $K$ as input. Note that, to reduce computational burden and as a proof of concept, we feed it a fixed-length prompt representation $h(x)$ extracted from the LLM forward pass, rather than the full embedded prompt. In our implementation, $h(x)\in\mathbb{R}^{2d}$ concatenates the mean-pooled first-layer and last-layer token embeddings, where each hidden-state sequence $H^{(\ell)}(x)\in\mathbb{R}^{T\times d}$ has token length $T$ and hidden size $d$:
\begin{equation}
h(x) = \Big[\ \mathrm{Mean}(H^{(0)}(x))\ ;\ \mathrm{Mean}(H^{(L)}(x))\ \Big].
\label{eq:h_prompt}
\end{equation}
Here, $H^{(0)}(x)$ and $H^{(L)}(x)$ denote the first- and last-layer hidden-state sequences for the prompt.

Still, as a proof of concept, architecture of \textsc{ScaleNet} is kept simple: a two-layer MLP followed by a non-negative output parameterization, since learning-rate scales must be non-negative. Concretely, for each LoRA matrix $\Delta W_i$ at step $k$, we map its unconstrained output $\alpha_i^{(k)}$ (optionally with a safety maximum clamp to avoid extreme values) for that block into a non-negative value via:
\begin{equation}
s_i^{(k)} \;=\; g\!\left(\alpha_i^{(k)}\right),\quad
g(a)=
\begin{cases}
\exp(a), & a \le 0,\\
1+a+\tfrac{1}{2}a^2, & a > 0.
\end{cases}
\label{eq:positive_map}
\end{equation}

We randomly draw $K$ uniformly from $\{0, 1,\dots,K_{\max}\}$ each time to support diverse TTA schedules and observe the trend. 

\paragraph{First-order Approximation.}
In our training framework, the hypernetwork is updated using a supervision loss computed \emph{after} running $K$ LoRA updates on LLM. Hence, optimizing hypernetwork parameters $\psi$ requires differentiating the post-adaptation answer loss $f(y;\Phi^{(K)})$ through $K$ unrolled TTA updates. One TTA update from step $k$ into step $k+1$ using prompt loss $f(x;\Phi^{(K)})$ can be expressed as:
\begin{equation}
\Phi^{(k+1)}=\Phi^{(k)}-\eta\, s^{(k)}(x,k,K;\psi)\,\nabla_{\Phi}f(x;\Phi^{(k)}).
\label{eq:unrolled_update}
\end{equation}
Differentiating Eq.~\eqref{eq:unrolled_update} yields \emph{second-order} dependency because
$\nabla_{\Phi}f(x;\Phi^{(k)})$ depends on $\psi$ again as LLM trainable parameters $\Phi^{(k)}(\psi)$ already depend on \textsc{ScaleNet} parameters $\psi$:
\begin{equation}
\frac{\partial}{\partial \psi}\nabla_{\Phi}f(x;\Phi^{(k)})
=
\nabla_{\Phi}^{2}f(x;\Phi^{(k)})\,
\frac{\partial \Phi^{(k)}}{\partial \psi}.
\label{eq:second_order_path}
\end{equation}
This second-order Hessian product is expensive and often unsupported by memory-efficient kernels used in modern LLMs (e.g., FlashAttention). So we instead adopt a first-order approximation by \emph{dropping} the second-order path in Eq.~\eqref{eq:second_order_path}, i.e., we treat prompt gradient $\nabla_{\Phi}f(x;\Phi^{(k)})$ as constant with respect to $\psi$. As a result, differentiating Eq.~\eqref{eq:unrolled_update} while ignoring $\tfrac{\partial}{\partial\psi}\nabla_{\Phi}f(x;\Phi^{(k)})$ yields:
\begin{equation}
\frac{\partial \Phi^{(k+1)}}{\partial \psi}
\approx
\frac{\partial \Phi^{(k)}}{\partial \psi}
-\eta\,\frac{\partial s^{(k)}(x,k,K;\psi)}{\partial \psi}\,
\nabla_{\Phi}f(x;\Phi^{(k)}).
\label{eq:fo_jacobian}
\end{equation}
Now gradients to $\psi$ flow only through the explicit dependence of the update magnitudes on $s^{(k)}(x,k,K;\psi)$, without requiring second-order derivatives.

\section{Experiments}
\subsection{Implementation Details}
We employ dynamic TTA and train \textsc{ScaleNet}, a two-layer MLP with hidden size 128, using AdamW at learning rate $10^{-4}$. The base TTA learning rate (to be scaled) is set to $10^{-2}$. For each dataset--model pair, we train on roughly 30k samples (mostly without repeats) and evaluate on 300 test samples. We set the LoRA rank to $r=4$ and LoRA $\alpha$ to 16. The maximum total number of TTA steps is $K_{\max}=5$. LoRA parameters are initialized with $B=0$ and $A\sim 10^{-2}\mathcal{N}(0,I)$, and are re-initialized randomly for each prompt. All LLM backbones use bfloat16.

\subsection{LLMs and Datasets}
We experiment with two families of LLMs: \textbf{Llama-3.2-3B}, \textbf{Llama-3.2-3B-Instruct} and \textbf{Llama3.3-70B-Instruct}; \textbf{Qwen3-4B}, \textbf{Qwen3-4B-Instruct} and \textbf{Qwen3-32B}. For each family, the ``Instruct" variant is post-trained for instruction-following and multi-turn dialogue. Evaluation is conducted across several datasets. \textbf{XSum} \cite{narayan2018dontdetailsjustsummary} is a summarization dataset of BBC news articles paired with single-sentence summaries. \textbf{SQuAD} \cite{rajpurkar2016squad100000questionsmachine} is a reading-comprehension benchmark where answers are spans from a provided Wikipedia passage. \textbf{NQ-Open} \cite{lee-etal-2019-latent} is an open-domain QA setting that evaluates predicting short answers to real user queries. \textbf{AdaptEval} \cite{hu2025testtimelearninglargelanguage} is a comprehensive benchmark for test-time learning that groups datasets into three categories: \emph{DomainBench} (Geography, Agriculture, Medicine, Finance), \emph{InstructionBench} (Alpaca-GPT4, Dolly, InstructionWild), and \emph{ReasoningBench} (GSM8K, MetaMath, LogiQA), covering diverse tasks and domains.

\subsection{Main Results}

\paragraph{Evaluation metric.}
Since we train with teacher-forced next-token negative log-likelihood (NLL), NLL is the most directly aligned evaluation metric. In practice, however, users often care more about generation-oriented metrics such as ROUGE-Lsum \cite{lin-2004-rouge}. Without dataset-level fine-tuning—which effectively teaches the LLM the desired answer distribution—reductions in NLL from unsupervised sample-specific TTA are not guaranteed to translate into consistent improvements in ROUGE-Lsum. This is justifiable because ROUGE-Lsum is based on lexical overlap, while natural language admits many valid paraphrases; the overlap score is therefore highly sensitive to surface form and stylistic choices, which are primarily learned through dataset-level supervision. Nevertheless, we report both NLL and ROUGE-Lsum.

\begin{table}[t]
  \caption{NLL results on \textbf{AdaptEval} for Llama3.3-70B-Instruct and Qwen3-32B under different TTA steps (lower is better).}
  \label{tab:large_adapteval}
  \centering
  \small
  \begin{tabular}{llccc}
    \toprule
    \textbf{LLM} & \textbf{Steps} & \textbf{Fixed} & \textbf{Step-wise} & \textbf{Layer-wise} \\
    \midrule
    \multirow{6}{*}{\shortstack[l]{Llama3.3\\-70B\\-Instruct}}
      & No TTA  & 2.2114  & 2.2114  & 2.2114 \\
      & 1 step  & 2.1907  & 2.1665  & \underline{\textbf{1.6598}} \\
      & 2 steps & 5.8488  & 2.1399  & \textbf{1.6692} \\
      & 3 steps & 9.6803  & 2.1850  & \textbf{1.6790} \\
      & 4 steps & 11.5891 & 2.1997  & \textbf{1.6741} \\
      & 5 steps & 11.4970 & 2.2657  & \textbf{1.7048} \\
    \midrule
    \multirow{6}{*}{\shortstack[l]{Qwen3\\-32B}}
      & No TTA  & 2.1805 & 2.1805 & 2.1805 \\
      & 1 step  & 2.1809 & 2.1303 & \textbf{2.1107} \\
      & 2 steps & 2.1784 & 2.0892 & \textbf{1.9168} \\
      & 3 steps & 2.1757 & 2.0842 & \textbf{1.8777} \\
      & 4 steps & 2.1725 & 2.0829 & \underline{\textbf{1.8739}} \\
      & 5 steps & 2.1682 & 2.0925 & \textbf{1.8889} \\
    \bottomrule
  \end{tabular}
\end{table}

\begin{table}[t]
  \caption{ROUGE-Lsum results (higher is better). Due to computation budget, we cut the maximum generated new token number at 64 for XSum, 32 for SQuAD and NQ, and 256 for AdaptEval.}
  \label{tab:rouge-lsum}
  \centering
  \small
  \begin{tabular}{llccc}
    \toprule
    \textbf{Pair} & \textbf{Steps} & \textbf{Fixed} & \textbf{Step-wise} & \textbf{Layer-wise} \\
    \midrule
    \multirow{3}{*}{\shortstack[l]{Llama3B\\-Instruct\\on XSum}}
      & No TTA  & 0.1821 & 0.1821 & 0.1821 \\
      & 1 step  & 0.1800 & 0.1838 & \textbf{0.1964} \\
      & 5 steps & 0.1780 & 0.1904 & \underline{\textbf{0.2090}} \\
    \midrule
    \multirow{3}{*}{\shortstack[l]{Qwen4B\\-Instruct\\on XSum}}
      & No TTA  & 0.1700 & 0.1700 & 0.1700 \\
      & 1 step  & 0.1700 & 0.1872 & \textbf{0.2157} \\
      & 5 steps & 0.1773 & 0.1864 & \underline{\textbf{0.2247}} \\
    \midrule
    \multirow{3}{*}{\shortstack[l]{Llama3B\\-Instruct\\on SQuAD}}
      & No TTA  & 0.7080 & 0.7080 & 0.7080 \\
      & 1 step  & 0.7058 & 0.7133 & \textbf{0.7238} \\
      & 5 steps & 0.6532 & 0.7114 & \underline{\textbf{0.7353}} \\
    \midrule
    \multirow{3}{*}{\shortstack[l]{Qwen4B\\-Instruct\\on SQuAD}}
      & No TTA  & 0.7722 & 0.7722 & 0.7722 \\
      & 1 step  & 0.7657 & 0.5361 & \textbf{0.7958} \\
      & 5 steps & 0.7634 & 0.5402 & \underline{\textbf{0.8306}} \\
    \midrule
    \multirow{3}{*}{\shortstack[l]{Llama3B\\-Instruct\\on NQ-Open}}
      & No TTA  & 0.2766 & 0.2766 & 0.2766 \\
      & 1 step  & \underline{\textbf{0.2722}} & 0.2662 & 0.2398 \\
      & 5 steps & 0.0288 & \textbf{0.2674} & 0.2507 \\
    \midrule
    \multirow{3}{*}{\shortstack[l]{Qwen4B\\-Instruct\\on NQ-Open}}
      & No TTA  & 0.1320 & 0.1320 & 0.1320 \\
      & 1 step  & 0.1424 & 0.0866 & \textbf{0.1513} \\
      & 5 steps & 0.1495 & 0.1041 & \underline{\textbf{0.1706}} \\
    \midrule
    \multirow{3}{*}{\shortstack[l]{Llama70B\\-Instruct\\on AdaptEval}}
      & No TTA  & 0.2327 & 0.2327 & 0.2327 \\
      & 1 step  & \textbf{0.2325} & 0.2068 & 0.2237 \\
      & 5 steps & 0.0029 & 0.2223 & \underline{\textbf{0.2733}} \\
    \midrule
    \multirow{3}{*}{\shortstack[l]{Qwen32B\\on AdaptEval}}
      & No TTA  & 0.0987 & 0.0987 & 0.0987 \\
      & 1 step  & \textbf{0.0992} & 0.0982 & 0.0983 \\
      & 5 steps & 0.1004 & 0.0990 & \underline{\textbf{0.1043}} \\
    \bottomrule
  \end{tabular}
  \vskip -0.1in
\end{table}

\paragraph{Analysis.}
We first report results in terms of NLL per answer token on a set of moderate-size LLMs: Llama-3.2-3B, Llama-3.2-3B-Instruct, Qwen3-4B, and Qwen3-4B-Instruct. As an empirical baseline, we run na\"ive TTA with a fixed learning rate of $5\times10^{-2}$ applied uniformly across all layers and adaptation steps. For our dynamically controlled TTA, we use a base learning rate of $10^{-2}$ and rescale it using the predicted multipliers. Because our protocol is sample-specific (memoryless), each prompt is permitted to contain the full task context; otherwise, the prompt-only objective can be ill-posed (see Appendix for detailed format).

To validate the effect of layer-wise control, we include a layer-agnostic/step-wise ablation that predicts a single multiplier per adaptation step and applies it uniformly across layers. \textbf{This step-wise variant is indispensable: it can be viewed as a learned upper bound over common hand-crafted learning-rate schedules} (e.g., linear decay, cosine annealing \cite{loshchilov2017sgdrstochasticgradientdescent}, exponential decay \cite{li2019exponentiallearningrateschedule}), since any such schedule can be expressed as a particular choice of per-step scalar multipliers. Remember that our layer-wise hypernetwork predicts per-step, per-layer multipliers (separately for Q/V projections), strictly generalizing step-wise control by enabling layer-specific modulation. Finally, we consider an additional ablation that averages the layer-wise scaling across samples to remove prompt dependence.

As shown in \cref{fig:base_LLMs}, the na\"ive baseline can rarely achieve optimal performance: it typically yields small initial gains, but then quickly degrades as NLL rises sharply with further steps, indicating destructive drift. In contrast, \textbf{our proposed method substantially improves stability and effectiveness}. Both the original layer-wise scaling and layer-agnostic/step-wise variants consistently outperform the fixed-rate baseline. \textbf{Introducing layer-wise control further reduces NLL, suggesting that different transformer blocks prefer different TTA update magnitudes}. Across datasets, most improvements occur in the first adaptation step, with subsequent steps providing diminishing returns. In addition, crudely averaging scales across samples can reduce the performance gains, indicating that \textbf{the dynamic scaling varies with prompt content}.

Overall, substantial variation is observed across dataset--model pairs. In particular, Llama-3.2-3B on AdaptEval shows very limited TTA gains. Since AdaptEval is more diverse and challenging—often requiring deeper logical reasoning—and to test whether our proposed unsupervised dynamic TTA framework scales to larger LLMs, we further evaluate Llama3.3-70B-Instruct and Qwen3-32B on AdaptEval. As listed in Table~\ref{tab:large_adapteval}, our dynamic TTA consistently outperforms the na\"ive baseline and the layer-agnostic/step-wise ablation on the 32B and 70B LLMs, with gains for Llama3.3-70B-Instruct even larger than for Llama-3.2-3B-Instruct on AdaptEval. \textbf{This implies our method has the potential to scale well into large-size LLMs}.

We next report ROUGE-Lsum results. For brevity, we report ROUGE-Lsum only at total step $K \in \{0,1,5\}$ for Llama-3.2-3B-Instruct and Qwen3-4B-Instruct on XSum, SQuAD, and NQ-Open, and for Llama-3.3-70B-Instruct and Qwen3-32B on AdaptEval. Table~\ref{tab:rouge-lsum} shows that our method can also improve ROUGE-Lsum, indicating that it is not simply a next-token NLL ``hacking'' trick, but can meaningfully improve generated text without requiring additional supervision or external material. The ROUGE-Lsum gains on XSum and SQuAD are clear and stable, while on NQ-Open and AdaptEval the changes are smaller and less stable. As a zero-shot setting, the former two datasets contain most of the answer information in the prompt itself, whereas the latter two are more open and often require harder reasoning, and are therefore more difficult to optimize using prompt-only adaptation \cite{ma2022opendomainquestionanswering}.

\begin{figure}
  \vskip 0.2in
  \begin{center}
  \vspace{-2mm}
    \centerline{\includegraphics[width=1.0\columnwidth]{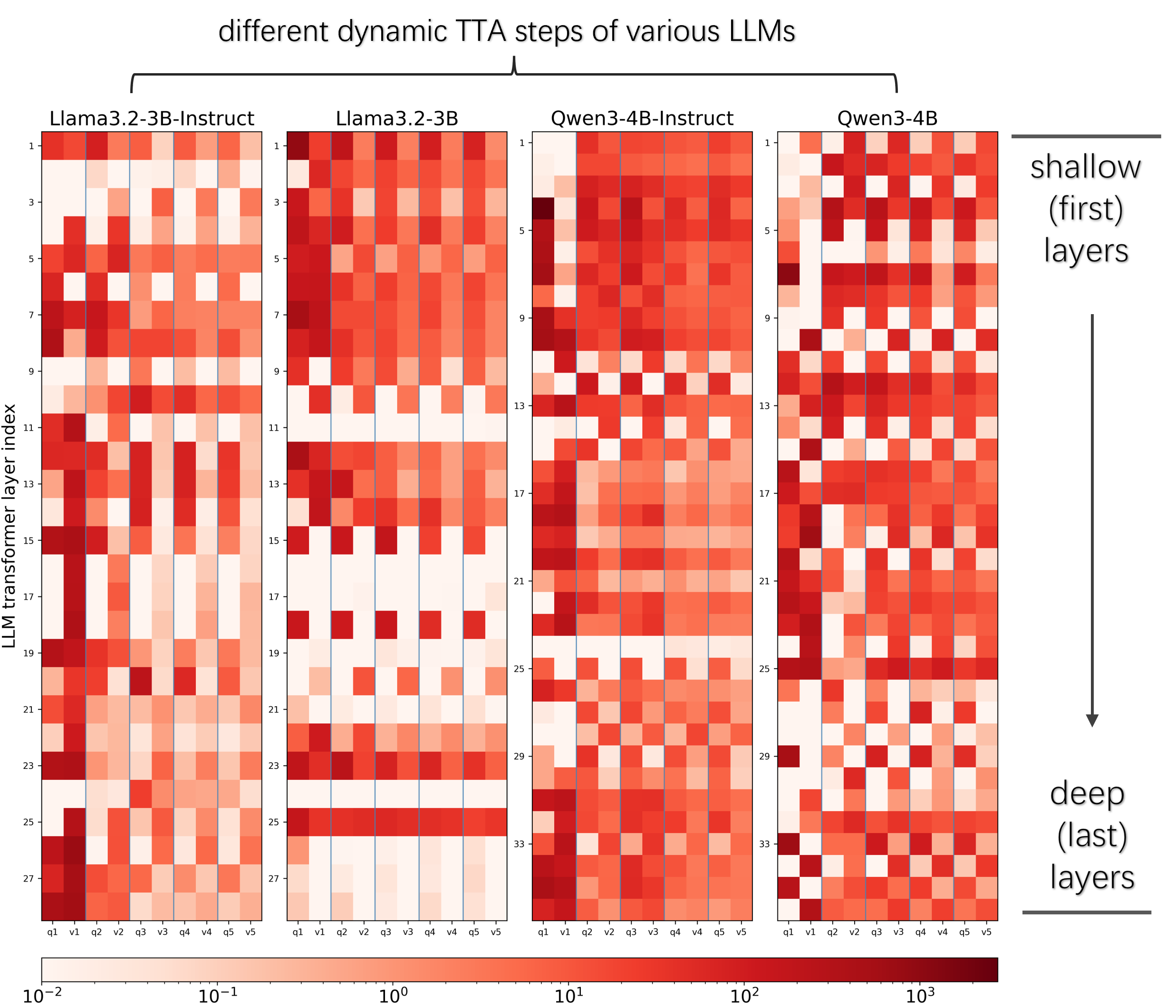}}
    \vspace{-2mm}
    \caption{
      \textsc{ScaleNet} output heatmap averaged over 4 datasets and 4 moderate-size LLMs. Along horizontal axis, $q_k$ and $v_k$ denote query and value projection at step $k$. From top to bottom, transformer layers are ordered from shallow (first) to deep (last).}
    \vspace{-2mm}
    \label{fig:heatMap}
    \vspace{-2mm}
  \end{center}
  \vspace{-2mm}
\end{figure}

\begin{figure}
  \vskip 0.2in
  \begin{center}
  \vspace{-2mm}
    \centerline{\includegraphics[width=0.95\columnwidth]{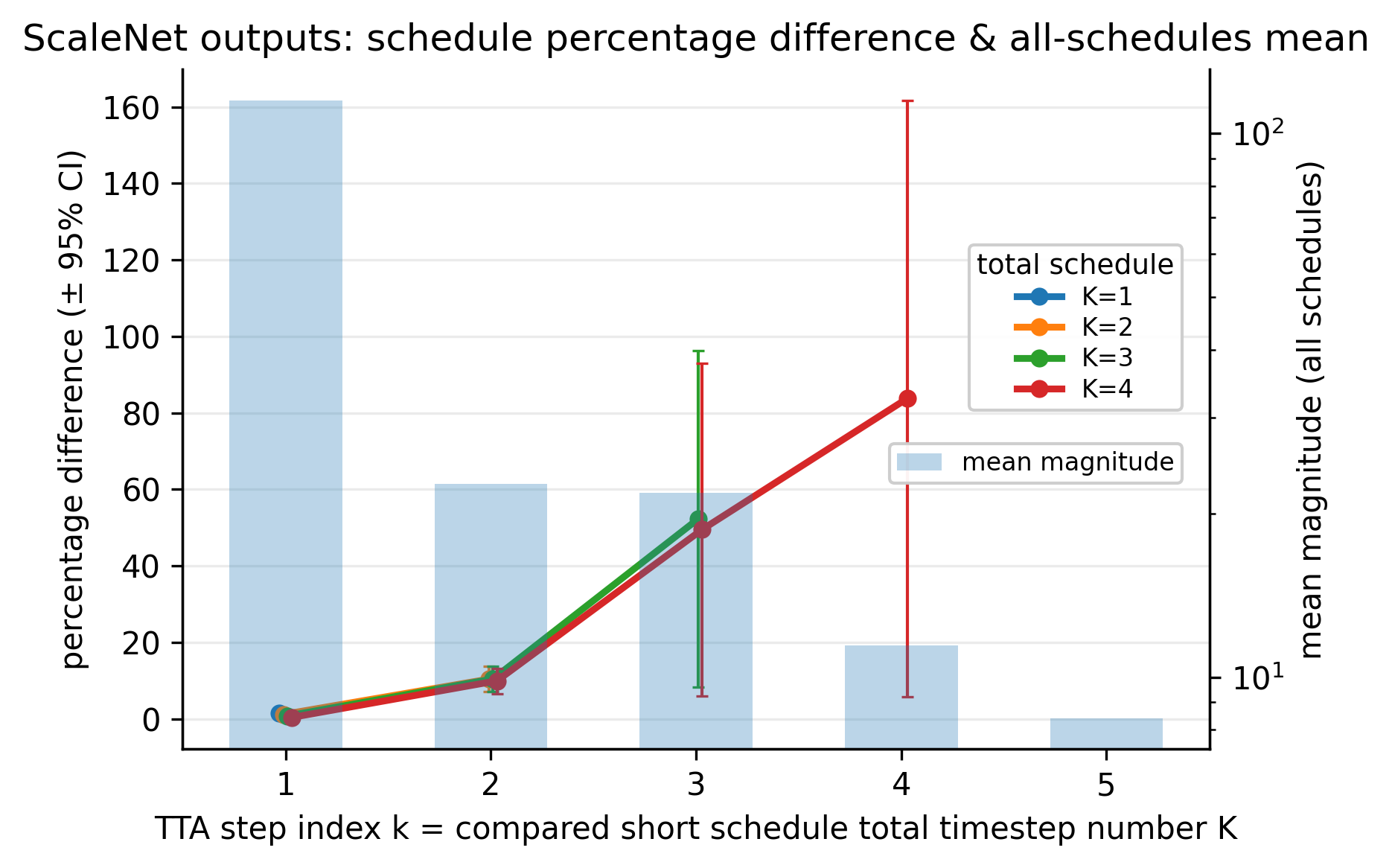}}
    \vspace{-2mm}
    \caption{
      \textsc{ScaleNet} output percentage difference with 95\% CI across schedules (baseline K=5) and all-schedules scaling magnitude mean. Both averaged over 4 dataset test samples and 4 moderate-size LLMs.}
    \vspace{-2mm}
    \label{fig:layer_factors}
    \vspace{-2mm}
  \end{center}
  \vspace{-2mm}
\end{figure}

\paragraph{Visualization.}
Here we visualize the learned learning rate multipliers. \Cref{fig:heatMap} shows that effective fine-grained dynamic learning-rate modulation has a rich structure, and that separating query/value projections matters. We do not observe a consistent monotonic trend along either the step dimension or the depth dimension (e.g., “query projection always larger than value projection,” or “early layers always smaller than late layers”). Instead, the preferred scales vary sharply across neighboring blocks: adjacent layers (or Q vs.\ V within the same layer) can differ by orders of magnitude even at the same TTA step. Notably, this heatmap is averaged over four datasets; the fact that high-contrast patterns remain after averaging suggests shared regularities rather than noise. For example, in Llama3.2-3B, several layers consistently favor much smaller update magnitudes, indicating that uniform learning-rate choices would over-update sensitive blocks. Together with \Cref{fig:layer_factors}, we observe that the average magnitude of \textsc{ScaleNet} outputs peaks at the first adaptation step ($k{=}1$) and then decays rapidly in subsequent steps, offering a natural explanation for why the majority of performance gains arise from the initial update.

Besides, for a stable TTA process, the update at a given step index $k$ should be comparable across schedules with different total steps $K$, since the per-step update rule is identical and only the horizon differs. As shown in \Cref{fig:layer_factors}, taking $K{=}5$ as the baseline, the average percentage differences (averaged over test samples) between scales from shorter schedules and the baseline at the same step index are relatively small. Given that the scales span several orders of magnitude, this indicates strong schedule consistency.

\subsection{Limitations}
Our current work prioritizes dataset-tailored, fine-grained, prompt-conditioned adaptation, which may limit transferability across substantially different task distributions. We believe this can be mitigated by training on larger corpus and using higher-capacity models beyond shallow MLP.

\section{Conclusion}
In summary, our proposed dynamic framework enables stable and effective unsupervised, sample-specific TTA by learning step- and layer-wise update magnitudes. It consistently outperforms the fixed-rate baseline and the step-wise variant, which represents common hand-crafted learning-rate schedules, suggesting that fine-grained, learned control is necessary.

\bibliography{example_paper}
\bibliographystyle{icml2026}

\newpage
\appendix
\onecolumn
\section{Prompt Formats for Unsupervised Sample-Specific TTA}
\label{app:prompt_format}

This section specifies the exact prompt templates used in our experiments. For each dataset, we construct a single text prompt $x$ and a gold answer $y$. During unsupervised sample-specific TTA, the model adapts \emph{only on the prompt tokens} (teacher-forced next-token NLL on $x$), and is then evaluated on the answer tokens (NLL on $y$ given $x$). For generation metrics (e.g., ROUGE-Lsum), we decode from the end-of-prompt cue (e.g., ``Summary:'', ``Answer:'', or ``\#\#\# Response:'') and compare the generated output to the gold answer.

\subsection{XSum (single-sentence news summarization)}
\label{app:xsum_prompt}
XSum pairs a news article with a single-sentence summary. We use the following template, ending with ``Summary:''.
\begin{quote}\small
\textbf{Prompt (XSum):}\\
Summarize the following news article in one concise sentence.\\
\\
Article:\\
\{\texttt{document}\}\\
\\
Summary:
\end{quote}
\noindent The gold target is $y=\{\texttt{summary}\}$.

\subsection{SQuAD (extractive reading comprehension)}
\label{app:squad_prompt}
SQuAD provides a passage and a question, and the answer is an exact text span from the passage. We explicitly instruct the model to output \emph{only} the span and end the prompt with ``Answer:''.
\begin{quote}\small
\textbf{Prompt (SQuAD):}\\
You are given a passage and a question.\\
Answer with the exact text span from the passage. Output ONLY the answer span.\\
\\
Passage:\\
\{\texttt{context}\}\\
\\
Question:\\
\{\texttt{question}\}\\
\\
Answer:
\end{quote}
\noindent The gold target is the first annotated answer span, $y=\{\texttt{answers}[0]\}$ (examples with empty answer lists are dropped).

\subsection{NQ-Open (open-domain short-answer QA)}
\label{app:nq_prompt}
NQ-Open contains a question and a short answer that may not appear in any given context. We use a minimal template that requires outputting only the answer string:
\begin{quote}\small
\textbf{Prompt (NQ-Open):}\\
Give a short answer of the following question. Output only the answer.\\
\\
Question: \{\texttt{question}\}\\
Answer:
\end{quote}

\subsection{AdaptEval (instruction / domain / reasoning benchmarks)}
\label{app:adapteval_prompt}
AdaptEval includes multiple instruction- and QA-style datasets. We convert each example into a unified \texttt{prompt}/\texttt{target} pair. All AdaptEval prompts end with ``\#\#\# Response:'' and the target is the reference response string.

\paragraph{Agriculture-QA (\texttt{agriculture-qa}).}
Each example contains a question and an answer string:
\begin{quote}\small
\textbf{Prompt (agriculture-qa):}\\
Below is an instruction that describes a task. Write a response that appropriately completes the request.\\
\\
\#\#\# Instruction:\\
\{\texttt{question}\}\\
\\
\#\#\# Response:
\end{quote}
\noindent The gold target is $y=\{\texttt{answers}\}$.

\paragraph{Reasoning sets with \texttt{instruction}/\texttt{output} (\texttt{MetaMathQA}, \texttt{gsm8k}, \texttt{logiqa}).}
We use \texttt{instruction} as the instruction text and \texttt{output} as the target:
\begin{quote}\small
\textbf{Prompt (MetaMathQA / gsm8k / logiqa):}\\
Below is an instruction that describes a task. Write a response that appropriately completes the request.\\
\\
\#\#\# Instruction:\\
\{\texttt{instruction}\}\\
\\
\#\#\# Response:
\end{quote}
\noindent The gold target is $y=\{\texttt{output}\}$.

\paragraph{General instruction-following schema with optional input (other AdaptEval subsets).}
For the remaining subsets, examples contain \texttt{instruction}, optional \texttt{input}, and \texttt{output}. If \texttt{input} is non-empty, we include an ``\#\#\# Input:'' block:
\begin{quote}\small
\textbf{Prompt (with input):}\\
Below is an instruction that describes a task, paired with an input that provides further context. Write a response that appropriately completes the request.\\
\\
\#\#\# Instruction:\\
\{\texttt{instruction}\}\\
\\
\#\#\# Input:\\
\{\texttt{input}\}\\
\\
\#\#\# Response:
\end{quote}

\begin{quote}\small
\textbf{Prompt (no input):}\\
Below is an instruction that describes a task. Write a response that appropriately completes the request.\\
\\
\#\#\# Instruction:\\
\{\texttt{instruction}\}\\
\\
\#\#\# Response:
\end{quote}
\noindent In both cases, the gold target is $y=\{\texttt{output}\}$.

You can have as much text here as you want. The main body must be at most $8$
pages long. For the final version, one more page can be added. If you want, you
can use an appendix like this one.

The $\mathtt{\backslash onecolumn}$ command above can be kept in place if you
prefer a one-column appendix, or can be removed if you prefer a two-column
appendix.  Apart from this possible change, the style (font size, spacing,
margins, page numbering, etc.) should be kept the same as the main body.

\end{document}